%% file: main.tex
\documentclass[runningheads]{llncs}

\usepackage[mobile]{eccv}

\usepackage{eccvabbrv}

\usepackage{graphicx}
\usepackage{booktabs}
\usepackage{indentfirst}
\usepackage{multirow}
\usepackage{multicol}
\usepackage{url}
\usepackage{enumitem}
\usepackage{makecell}
\usepackage[accsupp]{axessibility}  

\usepackage[colorlinks=false, hyperfootnotes=false]{hyperref}

\usepackage{orcidlink}

\begin{document}

\title{Attention Beats Linear for Fast Implicit Neural Representation Generation} 

\titlerunning{Attention Beats Linear for Fast Implicit Neural Representation Generation}

\author{
Shuyi Zhang$^{1,\dagger}$, 
Ke Liu$^{1,\dagger}$, 
Jingjun Gu$^{1,\dagger}$,
Xiaoxu Cai$^1$,
Zhihua Wang$^1$, \\
Jiajun Bu$^1$, 
Haishuai Wang$^{1,2,*}$ \\
\institute{
$^1$Zhejiang Provincial Key Laboratory of Service Robot, \\ College of Computer Science, Zhejiang University, Hangzhou, China \\ 
$^2$Shanghai Artificial Intelligence Laboratory, Shanghai, China\\
\email{
\texttt{\{zhangshuyi, keliu99, gjj, xiaoxu.cai, zhihua\_wang\}@zju.edu.cn} \\
\texttt{\{bjj, haishuai.wang\}@zju.edu.cn}
}}}

\authorrunning{S Zhang, K Liu et al.}

\maketitle

\begin{abstract}
Implicit Neural Representation (INR) has gained increasing popularity as a data representation method, serving as a prerequisite for innovative generation models. Unlike gradient-based methods, which exhibit lower efficiency in inference, the adoption of hyper-network for generating parameters in Multi-Layer Perceptrons (MLP), responsible for executing INR functions, has surfaced as a promising and efficient alternative.  However, as a global continuous function, MLP is challenging in modeling highly discontinuous signals, resulting in slow convergence during the training phase and inaccurate reconstruction performance. Moreover, MLP requires massive representation parameters, which implies inefficiencies in data representation. In this paper, we propose a novel Attention-based Localized INR (ANR) composed of a localized attention layer (LAL) and a global MLP that integrates coordinate features with data features and converts them to meaningful outputs. Subsequently, we design an instance representation framework that delivers a transformer-like hyper-network to represent data instances as a compact representation vector. With instance-specific representation vector and instance-agnostic ANR parameters, the target signals are well reconstructed as a continuous function. We further address aliasing artifacts with variational coordinates when obtaining the super-resolution inference results. 
Extensive experimentation across four datasets showcases the notable efficacy of our ANR method, e.g. enhancing the PSNR value from 37.95dB to 47.25dB on the CelebA dataset. Code is released at {https://github.com/Roninton/ANR}.
\end{abstract}

\footnote{$\dagger$ The first three authors contributed equally to this work. \\ * Corresponding author: Haishuai Wang (haishuai.wang@zju.edu.cn).}

\section{Introduction}
\label{sec:intro}

Implicit Neural Representation (INR) has recently demonstrated its proficiency in data representation.
Due to its adeptness in handling multi-modal data and its efficiency in representing high-dimensional data, INR serves as a foundational prerequisite for innovative generation models \cite{jun2023shap, chan2021pi}.
Specifically, INR can represent data such as 2D images and 3D scenes by the parameters of neural networks, i.e. Multi-layer Perceptrons (MLPs), that establish mappings from coordinate queries to data contents. 
Gradient-based methods \cite{sitzmann2020implicit, dupont2022data} have been employed to acquire the INR function by optimizing it through gradient descent to fit the target signals.
Although these methods could adeptly represent a data instance, their drawback lies in their inference inefficiency, as each inference necessitates additional gradient descent steps.
To overcome this limitation, Transformer-like hyper-networks \cite{chen2022transformers, kim2023generalizable} have been proposed to directly generate the parameters of MLP, which forms the INR function.
Applying a hyper-network simplifies the training process and benefits the downstream tasks, resolving the inference difficulty and enabling the model to learn a better inductive bias for the whole dataset. 

\begin{figure}[!b]
  \centering
\includegraphics[width=0.8\linewidth]{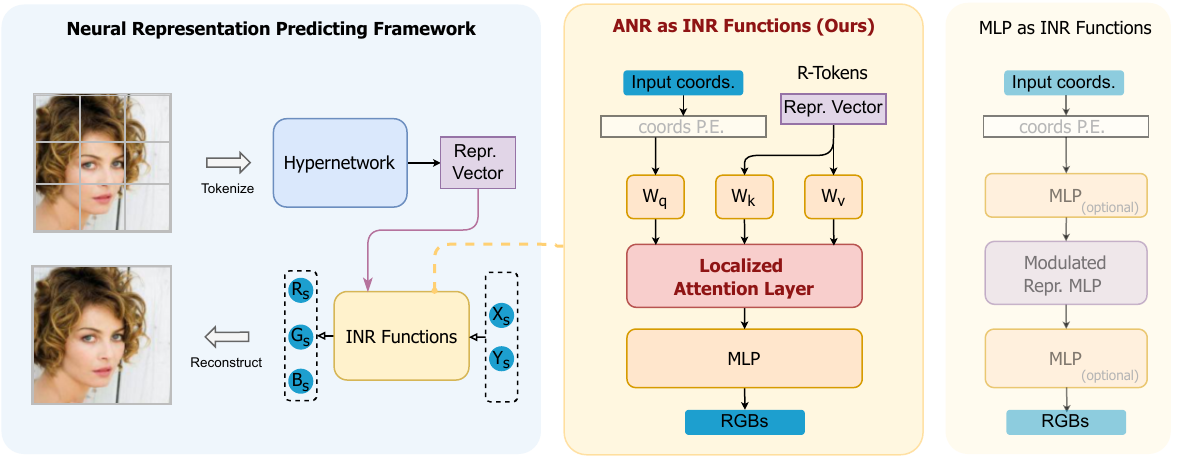}
   \caption{The framework overview and different INR functions: \textbf{(i) (Left)} The framework overview, INR reconstructs the mapping from coordinates and values.
   \textbf{(ii)(Mid)} ANR as INR functions. An ANR's parameters are data agnostic and it consists of a localized attention layer and linear components. We transform the coordinates inputs to attention Queries and get attention Keys/Values based on the previously generated R-Tokens. 
   \textbf{(iii)(Right) } MLP as INR functions. Except for the modulated MLP for representation data instance, the other MLPs are data agnostic.}
    \label{fig:anr_cover}
\end{figure}

\par

However, the MLP-based INR requires a substantial representation parameter size.
To mitigate the parameter explosion issue as the depth of MLPs increases, previous studies have attempted to modulate the representation parameters instead of the entire MLP. However, the effectiveness of modulation directly impacts the representation quality, thereby influencing downstream tasks.
\par
Moreover, the MLP-based INR fails to represent the details due to the exponentially increasing difficulty in learning high-frequency components \cite{liu2023partition}, such as edges in images, hairs in 3D meshes, and high-pitched sounds in audio data. 
Previous research has proposed methods to enhance MLPs for capturing high-frequency components, including the application of coordinate positional embedding (P.E.) \cite{tancik2020fourier, zheng2021rethinking, mildenhall2021nerf} and the introduction of periodic activation \cite{sitzmann2020implicit}. 
However, the integration of periodic activation is unsuitable for directly predicting INR with hyper-network due to its strict initialization scheme. The utilization of coordinate P.E. may give rise to complications, such as aliasing artifacts \cite{yuce2022structured}. 

\par
\begin{figure}[t]
  \centering
   \includegraphics[width=0.7\linewidth]{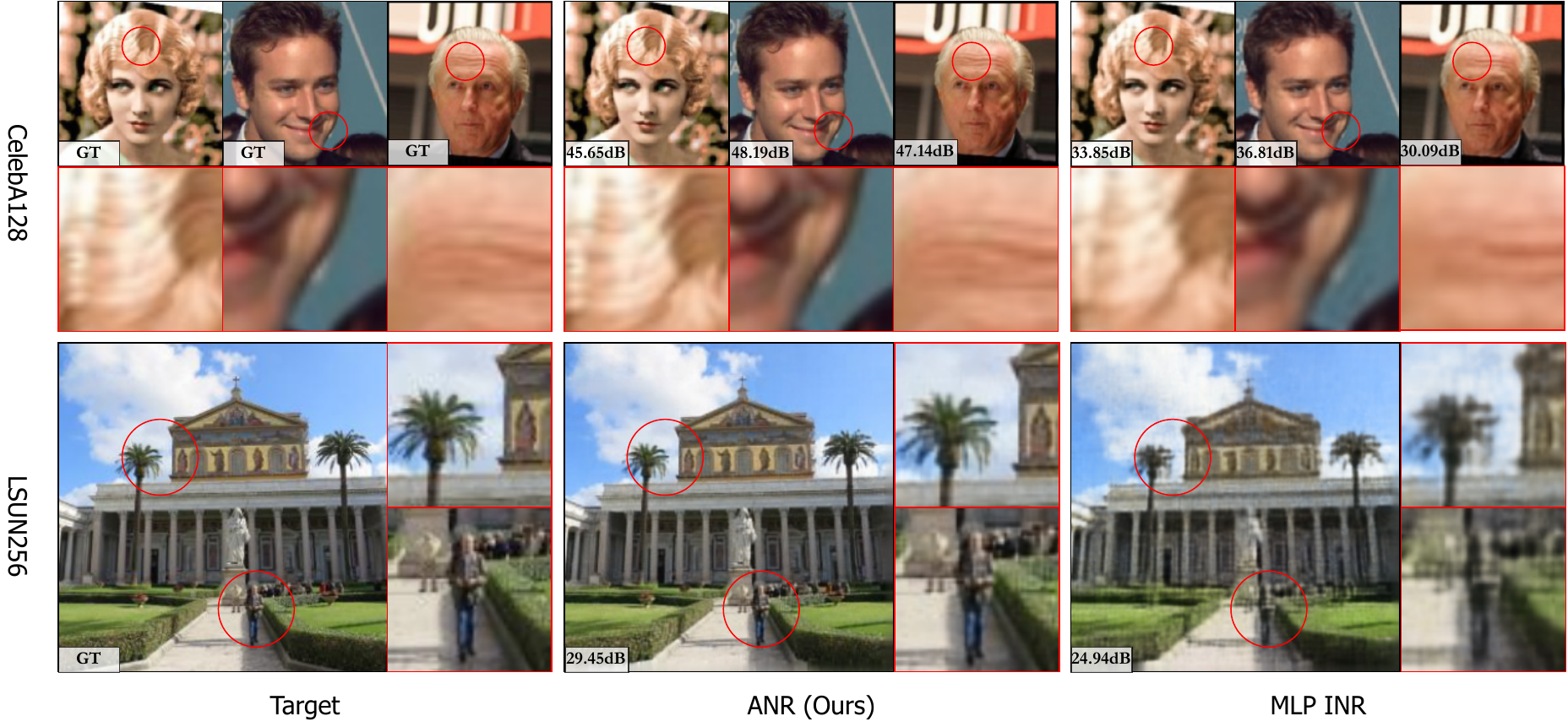}
   \caption{Reconstruction result of ANR and MLP-based INR, on CelebA ($128\times128$) and LSUN dataset ($256\times256$). Both experiments employ a training batch size 24 and are trained over 100k epochs. The red border images are the details of the original picture. }
   \label{fig:rec_res}
\end{figure}

In this paper, we integrate a localized attention layer (LAL) into INRs and propose a novel INR architecture, named \textit{\textbf{A}ttention-based localized implicit \textbf{N}eural \textbf{R}epresentation} (ANR), as illustrated in \cref{fig:anr_cover}. 
Our ANR accepts an instance-specific data feature and an instance-agnostic coordinate feature and applies an attention module to fuse them, followed by an instance-agnostic MLP that derives the output values.
To further improve the precision of the ANR, we introduce the L-softmax alignment in the attention fusion module, which ensures that gradients are backpropagated only through partial representation vectors.
Then, to acquire the feature of each data instance, we employ a transformer-like hyper-network to predict the instance-specific feature vectors, supplanting the conventional prediction of a huge amount of parameters in MLP-based INR \cite{chen2022transformers}.  
Furthermore, we address the INR aliasing problem with variational coordinates. 
The use of ANRs results in accelerated convergence and superior representation performance compared to the MLP-based INRs, as depicted in \cref{fig:rec_res}.
In summary, the key contributions of this work are as follows:
\par
\begin{itemize}[left=10pt,itemsep=2pt,topsep=0pt,parsep=0pt]
    \item We propose a novel attention-based INR architecture named ANR and a corresponding hyper-network to generate ANRs, which form continuous functions and efficiently represent large-size and high-frequency signals.
    \item We introduce a localized softmax alignment to focus ANR on the most related tokens, leading to better representation ability.
    \item We analyze the aliasing problem when generating INRs with the hyper-network and address the problem with variational coordinates, by examining the ANR mechanism and evaluating its characteristics. 
    \item Extensive experiments show that (i) ANRs could perform faster convergence and better representation than MLP-based INRs, and (ii) ANRs could handle various modalities of data, including images and  3D shapes.
    
\end{itemize}

\section{Related Work}
\label{sec:related_work}

\subsection{Implicit Neural Representation}
\label{sec:inr}

Implicit neural representation~\cite{sitzmann2020implicit} constructs a continuous mapping from input spatial or spatial-temporal coordinates $c \in \mathbb{R}^C$ to target signals $s \in \mathbb{R}^S$, where $C$ and $S$ are the dimensions of input and output respectively. Specifically, the neural network $f_\theta: \mathbb{R}^C \rightarrow \mathbb{R}^S$ can be regarded as a parametric representation of the data, where $\theta$ denotes the parameters of the network. Due to the efficiency of representing high-dimensional data \cite{dupont2022data}, INRs have worked well in tasks such as 3D shape representation \cite{zheng2022imface, chen2021neural} that may require massive storage space and computing power if using discrete data representations such as voxels and meshes. The target signal type could be just image RGB colors or other kinds of data attributes such as spatial transparency $\sigma$ that is often required in the Neural Radiance Field (NeRF) \cite{mildenhall2021nerf, lin2021barf}. 
Extensive research has shown the surprising potential of INR for various types of tasks, including graph information processing \cite{xia2023implicit}, vision data compression \cite{strumpler2022implicit}, super-resolution \cite{chen2022videoinr, zhang2022implicit}, surface representation \cite{zheng2022imface, chen2021neural, liu2023implicit}, volume rendering \cite{mildenhall2021nerf, anciukevivcius2023renderdiffusion, yuan2022nerf}, generation tasks \cite{jun2023shap, chan2021pi}, and so on. 
Interestingly, cooperating with global data agnostic modules, single-layer structure representation could just be enough to represent a data instance \cite{kim2023generalizable, benbarka2022seeing}. 
\par
As described before in \cref{sec:intro}, MLP-based INRs are a type of continuous representation used to store data information, regardless of spatial resolution. This approach has proven memory efficient, especially when dealing with high-dimensional data, such as 3D shapes. Conventionally, shape representation has been composed of voxel grids or triangular meshes. Although MLP-based INRs can theoretically fit any complicated mapping by making the network deeper and wider, they prioritize learning low-frequency signal components~\cite{liu2023partition, rahaman2019spectral}. Besides positional embedding and periodic activation, a few approaches have been innovated to speed up the convergence and rendering process. For example, decomposition-based methods~\cite{reiser2021kilonerf, liu2023partition} applied many tiny MLPs instead of one as INR functions, and Martel \etal \cite{martel2021acorn} utilized a multiscale adaptive coordinate network to achieve better results. The assumption of representation spatial proximity also benefits the representation  \cite{bauer2023spatial, lee2024locality}.
Although those methods do improve the representing ability of models, they either have usage limitations or introduce additional assumptions and computations.
Our method fuses localized attention within the representation and enables a more localized parameterization. This approach enhances the capability of INR functions rather than outside-representation improvement.
\subsection{Aliasing Problem}
\label{sec:inr_aliasing}
Y{\"u}ce \etal \cite{yuce2022structured} discussed the failing cases when applying implicit neural representation to fit specific target signals, including imperfect recovery and the aliasing problem. We find that MLP-based INRs generated suffer from the same situation. So, how to settle the aliasing problem could be a common challenge of MLP-based INRs \cite{landgraf2022pins, zhuang2022filtering}, and ANRs.
\par
Inspired by variational autoencoder \cite{kingma2013auto}, we summarized the reasons for the aliasing problem. For experimental convenience and better performance, we often use slightly higher or even excessively high positional embedding frequencies, introducing very high-frequency components into the input signals. 
However, during training, the sparsely sampled input query coordinates are fixed at specific positions. 
Subsequently, constraint points exist only on fixed sparse coordinates, while the frequency of sampled coordinates is much lower than those of high-frequency components.
Finally, the network learns a false amplitude for the excessively high-frequency input signals due to the low sampling frequency, which means the network learns false signal characteristics.
\par
Therefore, the aliasing problem means that INRs learn to reconstruct the signal with aliased high-frequency components, preventing INRs from reconstructing continuous and stable mapping.
\subsection{Attention Mechanism}
\label{sec:attn_mechanism}
The attention mechanism was initially proposed in computer vision \cite{larochelle2010learning, mnih2014recurrent} to reduce the computational complexity by focusing on specific parts of an image rather than the entire image. The introduction of the Transformer \cite{vaswani2017attention} demonstrated that the attention mechanism also shows great potential for other domains such as Natural Language Processing (NLP). Compared to sequence-based neural networks \cite{wang2022predicting,zhao2019attention}, architectures based on attention exhibited better parallelism. In subsequent research, models based on attention mechanisms found applications in a wide range of fields, including machine translation \cite{zhao2018attention}, image classification \cite{yang2016hierarchical}, image generation \cite{xu2015show}, audio processing \cite{zhao2019automatic}, and specific applications involving multi-modal data \cite{shen2018disan}.
\par
Brauwers \etal \cite{brauwers2021general} categorizes models using attention mechanisms into four components: query model, feature model, attention model, and output model. 
Researchers have extensively explored each of these components, such as applying hierarchical attention \cite{lu2016hierarchical} in the feature model and enhanced scoring and alignment methods \cite{ma2017interactive,luong2015effective} to address specific problems. 
Rebain \etal \cite{rebain2022attention} utilized cross-attention as a better alternative to feature concatenation for conditioning neural field, while still considering attention an outside-representation operation.
Our method modifies simple softmax alignment to introduce locality and make attention inside the representation structure.
\section{Method}
\label{sec:method}
We propose a novel implicit neural representation called \textbf{A}ttention-based Localized implicit \textbf{N}eural \textbf{R}epresentation (ANR) to address the convergence and inefficiency challenges of MLP-based INR for high-frequency data. 
As depicted in \cref{fig:anr_cover}, a hyper-network is designed to predict the feature vectors of data, also referred to as representation tokens, which we'll conveniently denote as `R-Tokens' in subsequent sections.
\par
R-Tokens is a feature matrix denoted as $D \in \mathbb{R}^{d \times N}$, where $N$ represents the number of feature tokens, and $d$ is the dimension of each feature. In the experiments, $N$ can be adjusted to accommodate varying data sizes. 
\par
Without loss of generality, we will focus on the problem of image signal reconstruction in this section. Specifically, we focus on fitting the mapping from query coordinates to the RGB values of the given images.
\subsection{Attention-based Localized INR}
\label{sec:ANR_arc}
\begin{figure*}[t]
  \centering
  \includegraphics[width=0.8\linewidth]{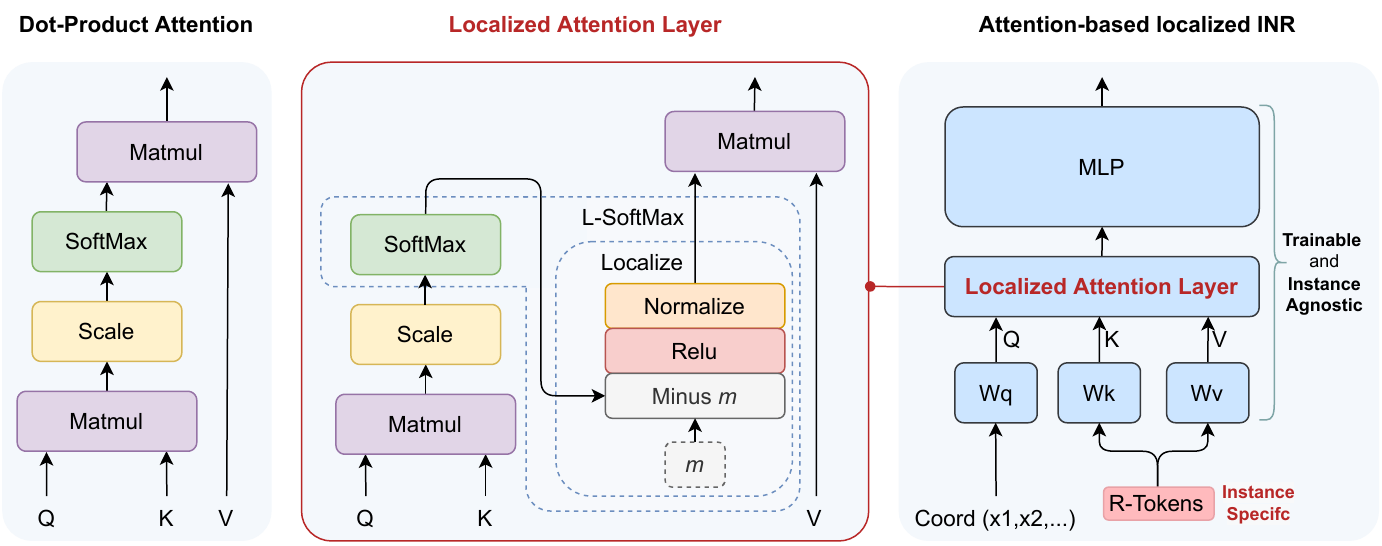}
   \caption{Visualization of Dot-Product Attention, Localized Attention Layer, and ANR. The term `m' is a hyper-parameter that serves to set a threshold for attention weights.}
   \label{fig:lal_anr}
\end{figure*}
\par
ANR is a novel INR architecture that benefits from the integration of the localized attention mechanism to the INR. Specifically, the input coordinate positional embeddings are considered query signals, and the Localized Attention Layer (LAL) can seek and filter the signals within the R-Tokens. The filtered signals are proven to be wave signals whose spectrum is expanded by the following MLPs.
Since R-Tokens are instance-specific and the following MLP is instance agnostic, we can treat the R-Tokens as modulated data representation and the remaining parts as a data-structure-related representation converter. 
From this view, we can consider ANR as the composition of an R-Tokens data representation and a representation converter.
\subsubsection{ANR Architecture.}
ANR simulates a mapping from the input coordinate $c$ to target signal $s$. Similar to normal INRs, it can be formulated as $f_\theta: \mathbb{R}^C \rightarrow \mathbb{R}^S$, where $C$ and $S$ are the dimensions of the input and output, respectively.
ANR consists of some necessary modules, and its details are depicted in \cref{fig:lal_anr}. Firstly, we apply Fourier mapping as the positional embedding (P.E.) operation, in which coordinate queries will be converted to a set of wave signals with different frequencies. 
Denoting $p$ as the dimension of $c$, the P.E. of $c$ should have the form:
\begin{equation}
\label{eqa:coord_pe}
\gamma(c) = sin(\Omega c + \phi),
\end{equation} 
where $\Omega \in \mathbb{R}^{p \times C}$ represents the P.E. spectrum and $\phi \in \mathbb{R}^{p}$ is the phase bias.
\par

Let $N$ be the token number of R-Tokens and $d$ their feature dimension. The matrix $D \in \mathbb{R}^{d \times N}$ represents R-Tokens. 
We transform R-Tokens to attention key and value matrices  $K, V \in \mathbb{R}^{d \times N}$ and $\gamma(c)$ to a query matrix $Q \in \mathbb{R}^d$ using linear projection. We formulate the process as below:
\begin{equation}
\label{eqa:coord_qkv}
Q = W_q \gamma(c), K = W_k D, V = W_v D,
\end{equation}
where $W_q \in \mathbb{R}^{d \times p}$, $W_k, \in \mathbb{R}^{d \times d}$ are transform matrices respectively. 
\par
Afterward, the Localized Attention Layer will fuse the features of D and $\gamma(c)$, and the subsequent MLP layers will expand the spectrum of the fused features and derive the final output. The localized Attention Layer will be introduced in the following subsection. 
Denoting ``$\circ$'' as the function composition operator, the forward process of ANR can be formulated as follows:
\begin{equation}
\label{eqa:anr_overrall}
f_\theta(c,D)=\texttt{ANR}(\gamma(c),D)= \texttt{MLP}\circ \texttt{LAL}(W_q\gamma(c) ,W_kD,W_vD).
\end{equation}
\subsubsection{Localized Attention Layer.}
\label{sec:lal_arc}
Compared to the origin attention mechanism shown in \cref{eqa:simple_attn}, we propose the localized attention mechanism with a more accurate attention alignment process. If we apply the softmax function to align attention weights (softmax alignment), every key holds a small weight in the attention map. Those small weights act as unrelated connections between queries and target signals, hindering the ANR's representation capability. We provide the comparison experiments in \cref{sec:threshold_in_lal}.
\par
As depicted in \cref{fig:lal_anr},  the LAL replaces the simple softmax with L-softmax. L-softmax utilizes a threshold $m$ as attention map activation to round those small attention map values to zero. The LAL can be formulated as \cref{eqa:L_attn}:
\begin{equation}
\label{eqa:simple_attn}
\texttt{Attention}(Q, K, V) = \texttt{softmax}\left(\frac{QK^T}{\sqrt{d_k}}\right)V,
\end{equation}
\begin{equation}
\label{eqa:L_attn}
\texttt{LAL}(Q, K, V) = \texttt{L-softmax}_{m}\left(\frac{QK^T}{\sqrt{d_k}}\right)V, 
\end{equation}
where $Q$, $K$, and $V$ stand for query, key, and value of the attention layer, $d_k$ stands for the length of the attention feature, and $m$ is a constant hyper-parameter used in the L-softmax alignment, respectively. 
\par
The L-softmax alignment strengthens the locality of the attention layer by selecting and combining only the tokens nearest to the query. The L-softmax makes the output localized and stops those irrelevant backpropagated gradients. We formulate the process as \cref{eqa:L-softmax}:
\begin{equation}
\label{eqa:L-softmax}
\texttt{L-softmax}_{m}(x) = \texttt{Localize}(\texttt{softmax}(x),m),
\end{equation}
where $x$ is the computed attention weights and the Localize operation normalizes the input matrices so that their last dimension sums up to one, as defined below:
\begin{equation}
\label{eqa:Localize}
\texttt{Localize}(x,m) = \texttt{Normalize}(\texttt{ReLU}(x-m)).
\end{equation} 
After applying the attention map activation, the irrelevant weights are removed, accelerating the network's convergence. 
\par

\subsubsection{The LAL Serves as a Signal Filter.}
\par
We delve into a detailed examination of the properties of the ANR and investigate its constituent modules. 
\par
In the initial input layer, each neuron $x$ is transformed utilizing Fourier position embedding. 
Specifically, each neuron is converted into a wave signal by a trigonometric function $\gamma_i(x) = \sin(\omega_i x + \phi_i)$ (also referred to as a wave function). 
At this stage, the input layer’s neurons consist of a single wave function, where $\omega_i$ is considered the base frequency, and $\phi_i$ denotes the phase bias. 
\par
We treat the Fourier embedding as initial \textbf{wave signals}. 
For those neurons whose output is obtained by summing wave functions with specific weights, we consider each of them to be \textbf{a set of wave signals} with spectrum $\Omega$.
Let $\Psi(x, \Omega )$  be a set of wave signals with the form:
\begin{equation}
    \Psi(x, \Omega ) =\sum_{\omega\in\Omega}A_{\omega}\sin\left(\langle\omega,x\rangle+\phi_{w}\right),
\end{equation}
where $x$ represents the spatial variable, $\omega$ is the basic frequency belongs to the frequency set $\Omega$, $A_{\omega}$ is the signal magnitude, and  $\phi_{\omega}$ is phase bias. 

\begin{proposition}\label{prop:prop1}
The application of multiplication and addition operations to a set of wave signals $\Psi(x, \Omega )$ results in a new set of wave signals $\Psi(x, \Omega^\prime)$.
\end{proposition}

\begin{proposition}\label{prop:prop2}
All analytic activation functions (e.g. ReLU and sinusoids) can be effectively approximated by polynomial functions $\rho(x) = \sum_{k=0}^{K}\alpha_kx^k$, thereby expanding the spectrum of the original wave signals $\Psi(x, \Omega )$ and generating a new set of waves $\Psi(x, \Omega^\prime )$.
\end{proposition}

These propositions, as demonstrated by Y{\"u}ce \etal \cite{yuce2022structured}, underpin our understanding of the Localized Attention Layer (LAL). 
As discussed in Section \ref{sec:lal_arc}, the LAL primarily encompasses: 

\begin{itemize}[left=25pt,itemsep=2pt,topsep=0pt,parsep=0pt]
    \item[(1)] Linear transformations, including scaling and matrix multiplication. 
    \item[(2)] Non-linear ReLU activation.
    \item[(3)] Filtering, normalization, and softmax activation.
\end{itemize}
\par
It's worth noting that the softmax activation can be viewed as a normalization step following an exponential function. Because the input to the softmax function is an attention weight that is much smaller than 1, we can represent the exponential function using the first $K$ terms of its Taylor series expansion:
$\text{exp}(x) = \sum_{k=0}^{K} \frac{x^k}{k!}$. 
\par
Drawing from \cref{prop:prop1} and \cref{prop:prop2}, we infer that operations (1, 2) within the LAL merely process the input wave signals to a new wave set.
So we can infer that the core mechanism of the LAL is the filtering and normalization (operation 3), which differentiates it from an MLP.
In conclusion, the LAL serves as a signal filter. We provide a more detailed discussion in \cref{sec:lal_disc_detail}.
\par
\subsection{Anti-aliasing with Variational Coordinates}
In our initial experiment, we observed that coordinate P.E. with low frequencies resulted in very smooth reconstructions, showing that the attention layer can fit continuous signals even if dot-product attention layers exhibit extreme nonlinearity \cite{kim2021lipschitz}. In other words, ANR is still a continuous signal representation. We provide the experiment details in \cref{sec:lowf_exp}.
\par
However, overly high-frequency inputs led to the aliasing problem as we discussed in \cref{sec:inr_aliasing}, which means that the interpolated points between original coordinates could be meaningless. It should be emphasized that the aliasing problem is widespread in all kinds of INR architectures. 
\begin{figure}[t]
  \centering
   \includegraphics[width=0.45\linewidth]{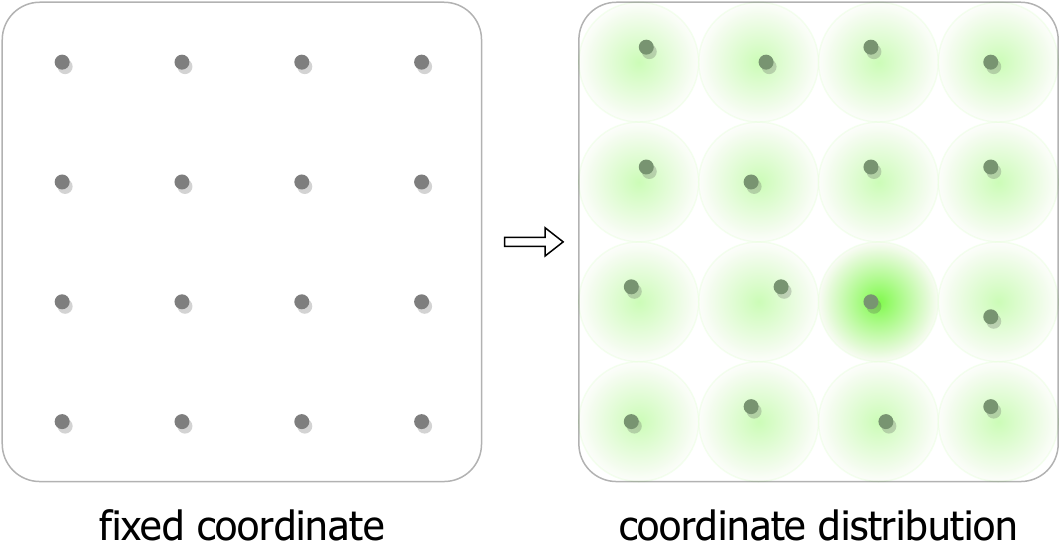}

   \caption{Visualization of the Variational Coordinates. Original coordinates are sampled at specific fixed positions. Our strategy is to sample coordinates from a distribution.}
\label{fig:noise_distribution}
\end{figure}
\par
Taking inspiration from variational autoencoders (VAEs) \cite{kingma2013auto}, we expand the coordinate sample space from fixed positions to a presumed distribution. 
We posit that coordinates, like any real-world data, are subject to noise. Therefore, we introduce Gaussian shifts to coordinates for better modeling of the inherent variability in empirical data, not just a regularization strategy.
\par
As illustrate in \cref{fig:noise_distribution}, we sample the variational coordinates of each pixel $c_{\rm var}$ that is perturbed based on uniform-sampled coordinates $c$, as shown below:
\begin{equation}
\label{eqa:coord}
c_{\rm var} = c + \alpha V, \quad V \sim \mathcal{N}(0, \sigma^2), \quad V \in [ - v_{\rm dev}, v_{\rm dev} ],
\end{equation}
\noindent where $\sigma$ represents the standard deviation, $\alpha$ is the scaling factor, and $v_{\rm dev}$ denotes a constraining factor to limit the maximum deviation. This constraint ensures that the fluctuation range of sampled coordinates falls within the range of adjacent original coordinates. 
\par
After resolving the aliasing problem, the interpolation in coordinate space becomes a meaningful and natural super-resolution method. To aid comprehension, we provide a brief visualization of how to solve the aliasing problem in 1D signal fitting in \cref{sec:1d_aliasing}.

\par
\section{Experiments}
\label{sec:experiment}

\par
In this section, we first provide an overview of the experimental setup and dataset collection. 
Subsequently, we present the experiment details on reconstruction tasks and view synthesis tasks 
on which we compare the convergence speed and parameter efficiency of ANR and MLP-based INR.
Lastly, we gave an ablation study about the ANR. 
\subsection{Experimental Settings}
\textbf{Data Collection}.
Our study utilizes a diverse set of datasets: CelebA \cite{liu2015faceattributes} for human facial images, LSUN \cite{yu2015lsun} for outdoor scenes, ERA5 \cite{hersbach2019era5} for global temperature data, and Learnit Shapenet's Cars and Chairs subsets \cite{chang2015shapenet} for view synthesis tasks. 
\par
\textbf{Baseline Comparisons}.
We benchmark our proposed ANR model against INR based on MLP \cite{chen2022transformers, kim2023generalizable} with different parameters modulation methods, e.g. grouping used by TransINR \cite{chen2022transformers} and matrix modulation used by IPC \cite{kim2023generalizable}. The modulation method enables the hyper-network to predict only a small number (the ``mod'' value) of columns instead of the whole transform matrix of MLP. All experiments are conducted under controlled conditions, keeping the hardware and software settings constant across different model iterations.
Moreover, we provide the experiment details of the periodic activation function's performance in \cref{sec:periodic_act}, which we found unsuitable for predicting INR utilizing a hyper-network framework.
\par
\textbf{Evaluation Metrics}.
We use mean squared error (MSE) as the loss function for all reconstruction tasks, which is consistent with typical reconstruction tasks. 
The evaluation metrics encompass the average PSNR computed on the validation set and the representation parameter size (the size of the feature vectors).

\textbf{Experimental Procedures}.
The model is trained using a batch size of 18 for the CelebA dataset, 16 for the LSUN dataset and ERA5 Dataset, and 18 for the Learnit Shapenet Dataset (sample 512 rays for each data instance), respectively. The training learning rate is 5e-5, and the training optimizer is Adam. The model is trained for 200k steps on CelebA, 100k steps on LSUN, 70k steps on ERA5, and 200k steps on Shapenet. The depth of both the encoders and decoders of the hyper-network is 6 for all experiments. 
\
To further improve the performance of ANR, we explore the hyper-network structure and find that a hyper-network that utilizes cross-attention is better than self-attention \cite{chen2022transformers}. We provide the discussion of the hyper-network structure in \cref{sec:trc}.

\textbf{Other Settings}.
It's worth noting that the \textbf{feature vector} for data representation is identified as R-Tokens when employing the hyper-network to predict R-Tokens for ANR. Conversely, if the hyper-network is utilized for predicting MLP-based INR, the feature vector comprises the weights of the MLP.
We provide more experiment implementation details in \cref{sec:impl_details}. 
\subsection{ANR Performance Overview}
\begin{figure*}[t]
\centering
\begin{minipage}{0.49\textwidth}
\centering
\includegraphics[width=\linewidth]{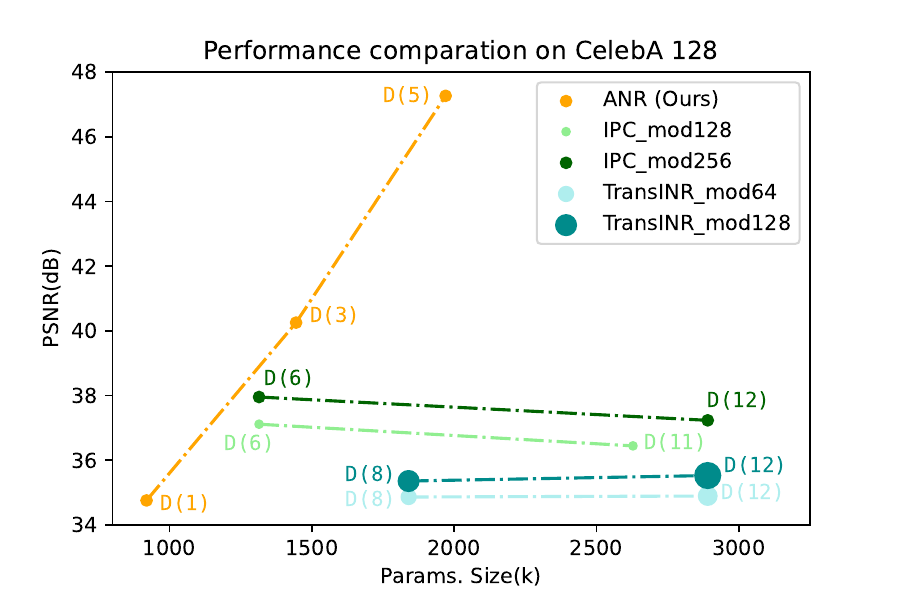}
\end{minipage}%
\begin{minipage}{0.49\textwidth}
\centering
\includegraphics[width=\linewidth]{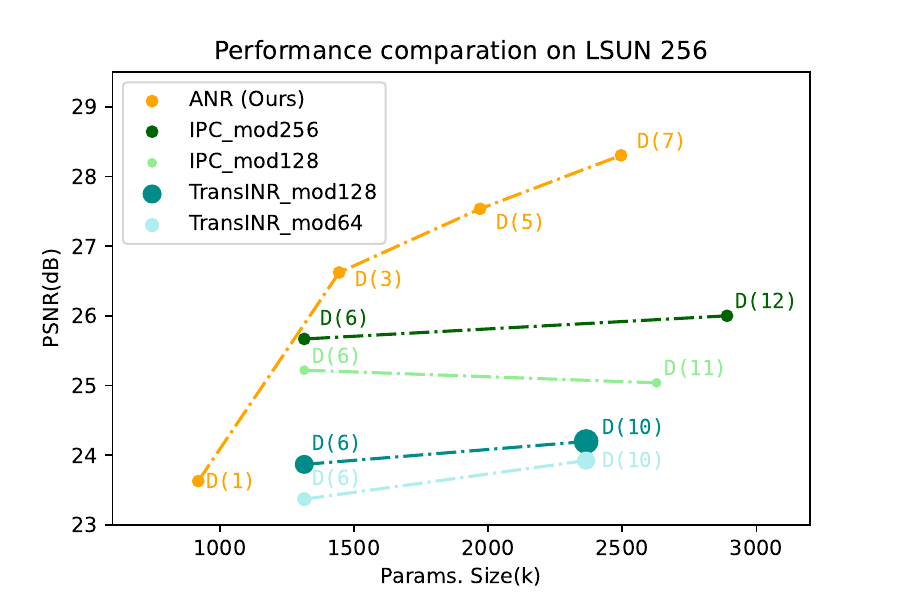}
\end{minipage}
\caption{Reconstruction performance on CelebA dataset and LSUN dataset. The size of each point stands for the representation sizes of data instances. The term ``\_mod(x)'' stands for the parameters modulation setting used by TransINR \cite{chen2022transformers} and IPC \cite{kim2023generalizable}, which limits the predicted maximum columns of MLP weights. The depth of the whole MLP used in the INR functions is annotated beside the data points with the form ``D(x)''. 
}
\label{fig:celeba_lsun_exps}
\end{figure*}
\begin{table*}[t]
  \centering
  \caption{Evaluations are conducted on image data, specifically CelebA dataset and LSUN dataset, as well as earth temperature fields obtained from ERA5. We compare our ANR with TransINR \cite{chen2022transformers} and IPC \cite{kim2023generalizable}. The term "Repr. Params" denotes the size of the feature vector generated by the hyper-network for data representation. The term "All Params" encompasses the total count of parameters for all functions within the Incremental Neural Representation (INR) framework.}
  \begin{tabular}{ccccccccc}
    \toprule
    Dataset& Input Size& Method& \makecell{Global/Repr. \\  MLP Depth}&MSE$\downarrow$ & PSNR$\uparrow$ & \makecell{Repr.\\Params}&  \makecell{All\\Params}& Steps \\
    \midrule[2pt]
     \multirow{4}*{CelebA}& \multirow{4}*{128*128*3}& TransINR & 0/12&27.5e-5& 36.25& 722k& 2891k&200k\\
    & & IPC & 11/1& 21.4e-5 & 37.23 & 131k & 2891k & 200k \\ 
    & & IPC & 5/1& 18.5e-5 & 37.95 & 131k & 1314k & 200k \\ 
    & & \textbf{ANR(Ours)} & 5/0&\textbf{2.2e-5}& \textbf{47.25}& \textbf{131k}& 1970k&200k\\
    \midrule
    \multirow{5}*{LSUN}& \multirow{5}*{256*256*3}& TransINR & 0/12&4.6e-3& 24.21& 591k& 2365k&100k\\
    & & IPC & 11/1& 3.1e-3 & 26.00 &  131k & 2891k & 100k \\ 
    & & IPC & 5/1& 3.3e-3 & 25.67 &  131k & 1314k & 100k \\ 
    & & ANR(Ours) & 5/0& 2.2e-3 & 27.54& 131k & 1969k&100k\\
    & & \textbf{ANR(Ours)} & 7/0&\textbf{1.9e-3}& \textbf{28.30}& \textbf{131k}& 2757k&100k\\
    \midrule
    \multirow{3}*{ERA5}& \multirow{3}*{181*360*1}& TransINR & 0/8 &38.5e-6& 44.65& 229k& 1839k&70k\\
    & & IPC & 7/1 & 40.8e-6 & 44.05 & 131k & 1839k & 70k \\
    & & \textbf{ANR(Ours)} & 5/0&\textbf{9.3e-6}& \textbf{51.49}& \textbf{131k}& 1969k&70k\\
    \bottomrule
  \end{tabular}
  \label{tab:performance_CLE}
\end{table*}

We show that ANRs lead to better convergence on reconstruction tasks, as depicted in \cref{tab:performance_CLE}. 
Our experiments on CelebA and LSUN show that ANRs require a smaller size of feature vector (Repr. Params) to represent each data instance, as shown in \cref{fig:celeba_lsun_exps}. 
Moreover, ANR can efficiently represent relatively invariant characteristics and perform better on the ERA5 dataset.  
We also test ANR for other tasks directly and find it performs well. The inference examples of ANR on Learnit Shapenet are shown in \cref{fig:volume_rendering} and those on ERA5 are shown in \cref{sec:inf_example_era5}.
We also provide additional results for model performance comparison in \cref{sec:more_res}.
In conclusion, ANR achieves \textbf{faster and more accurate convergence} and exceeds in \textbf{data representation efficiency}. 
\par 
\subsection{Attention Beats Linear on Reconstruction Tasks}
 \textbf{CelebA}. We first compared the convergence speed and effectiveness of ANR and MLP-based INR on the human facial dataset CelebA. Every input image is scaled to $128\times128$ and then normalized to $[0,1]$. Images in the CelebA dataset hold many edges, such as hairs, eyes, and various backgrounds. Experiments show that ANR outperforms MLP-based INR significantly on this dataset. 
\par
\textbf{LSUN}. The performance of ANR and MLP-based INR are also compared on the LSUN church dataset. We set the size of the target images to $256\times256$ so that images contain more details. Experiments show that on the LSUN dataset, ANR still exhibits significantly faster convergence. 
\par
\textbf{ERA5}. Although the input coordinates of the Earth's surface temperature data are in latitude and longitude, we can consider it as a numerical field with Cartesian coordinates. 
However, because the earth's temperature field is highly dependent on the earth's physical structures, the target image always holds similar borders. Experiments show that ANR can better represent these less varied data and it is more parameter-efficient.
\subsection{ANR is Efficient for View Synthesis} 
To demonstrate that the memorization of the radiance field by ANR is exclusively achieved through R-Tokens, we employ ANR directly to model the radiance field, ensuring the MLP representation converter remains data-agnostic. Experiments show that ANR performs view synthesis like NeRF \cite{mildenhall2021nerf} by establishing a mapping from spatial coordinates $(x,y,z)$ to opacity $\sigma$ and color (RGB), simulating the physical propagation of light rays in the real world. 
\par
\begin{figure}[t]
  \centering
   \includegraphics[width=0.6\linewidth]{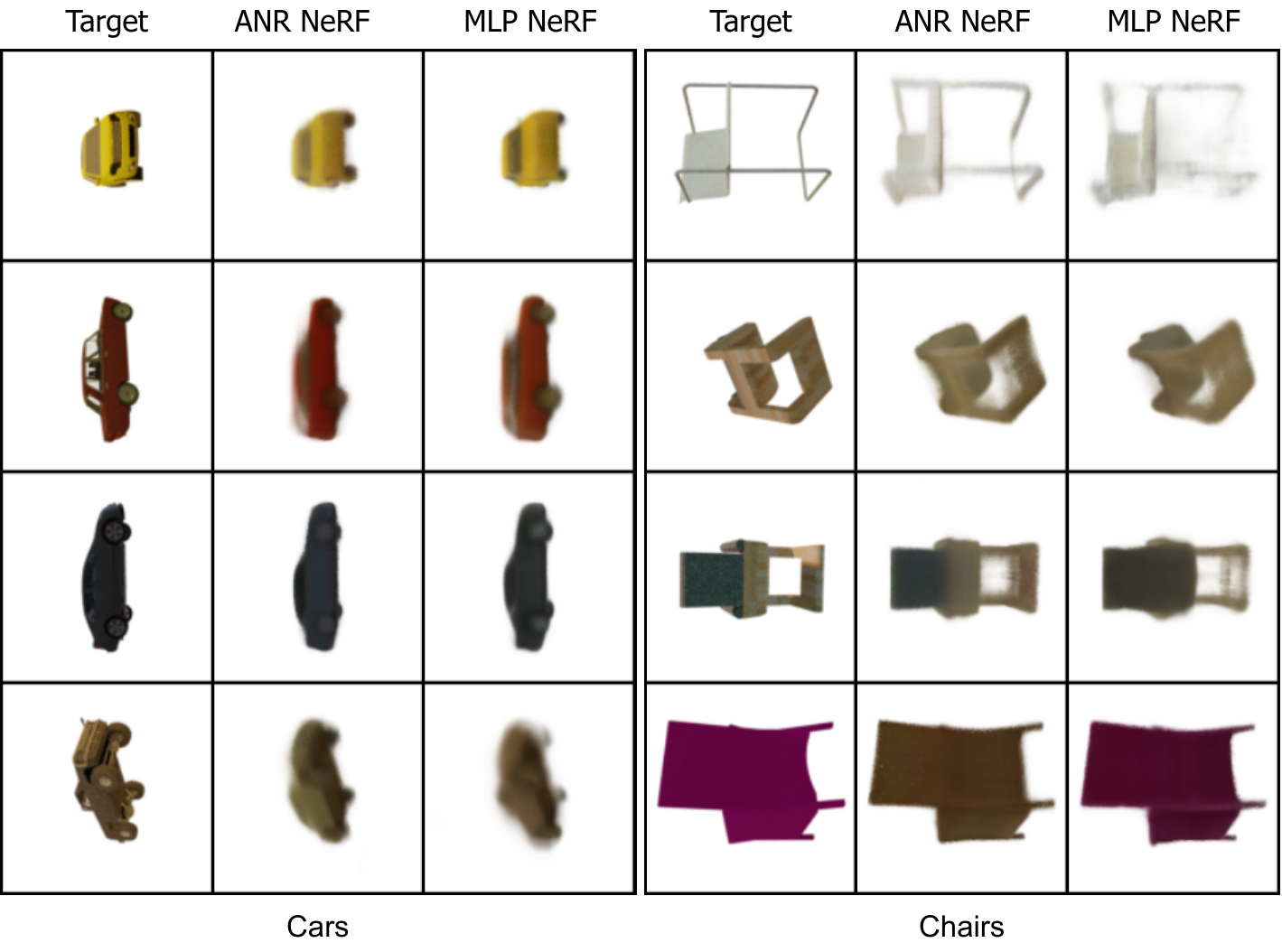}
   \caption{The view synthesis result of NeRFs on the Learnit Shapenet dataset with the same training step setting. ANR NeRFs memorize shape information better, while MLP NeRFs tend to memorize mean shapes and mean colors first.}
   \label{fig:volume_rendering}
\end{figure}

\textbf{Learnit Shapenet}. We conduct experiments on Learnit Shapenet's subsets Cars and Chairs. The view synthesis results are depicted in \cref{fig:volume_rendering}, showing that NeRFs based on ANR memorize shape information better, while those based on MLP tend to memorize mean shapes and mean colors first. We evaluate the mean PSNR values on the Cars and Chairs test set separately. The experiments indicate that the PSNR performance of ANR NeRF is better than IPC NeRF, as shown in \cref{tab:performance_ls}. 
\begin{table}[t]
  \centering
  \caption{Results of view synthesis on Learnit Shapenet Cars and Chairs. We compare our ANR with IPC \cite{kim2023generalizable}. The term "Repr. Params" refers to the feature vector size. The experiments are trained over 100k steps.}
  \begin{tabular}{cccccc}
    \toprule
       Shape&NeRF&MSE$\downarrow$ & PSNR$\uparrow$ & Repr. Params &Steps\\
    \midrule[2pt]
    \multirow{2}*{Cars}&IPC&3.59e-3& 26.44& 131k &200k\\
      &\textbf{ANR(Ours)}&\textbf{3.45e-3}& \textbf{26.57}& \textbf{131k} &200k\\
    \midrule
       \multirow{2}*{Chairs}&IPC&12.2e-3& 21.07& 131k &200k\\
      &\textbf{ANR(Ours)}&\textbf{11.0e-3}& \textbf{21.45}& \textbf{131k} &200k\\ 
    \midrule
  \end{tabular}
  \label{tab:performance_ls}
\end{table}

\subsection{Ablation Study}
    
\begin{figure}[t]
  \centering
   \includegraphics[width=0.55\linewidth]{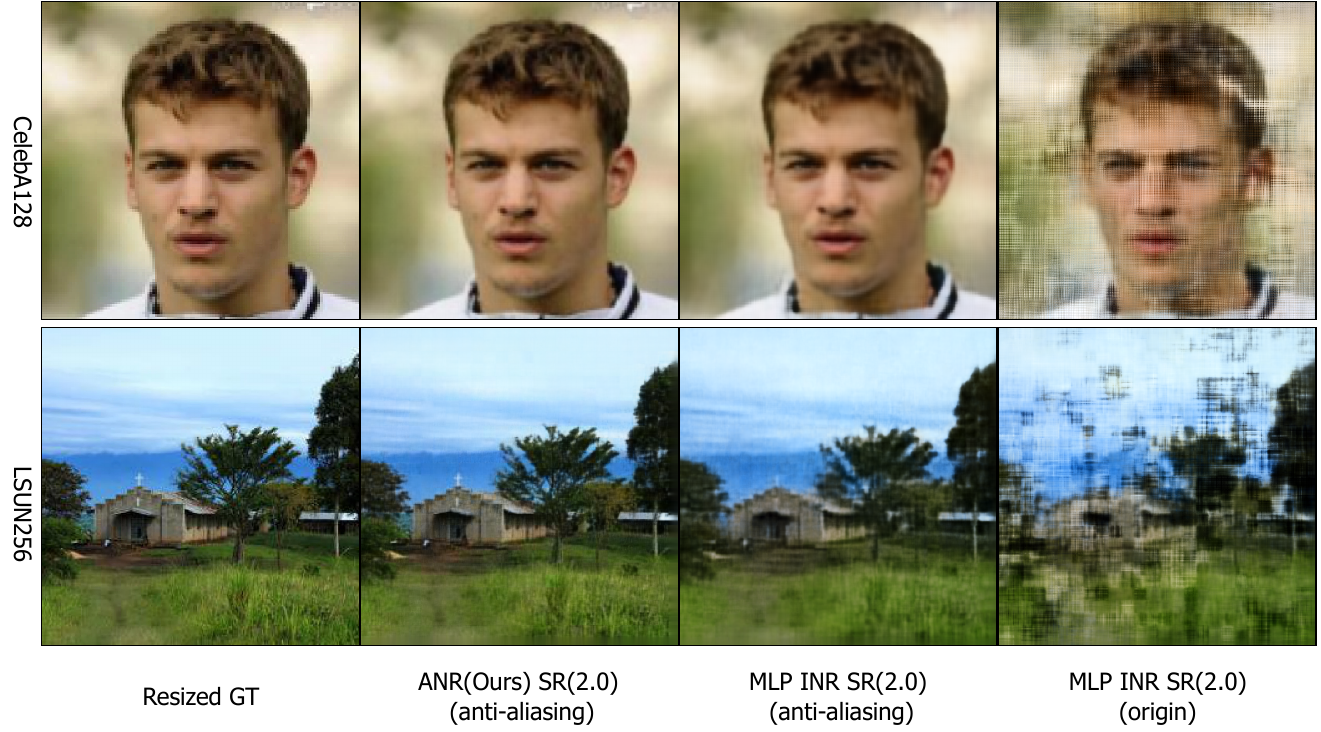}
   \caption{Super-resolution result of ANR and MLP-based INR. SR scaling ratio is 2.0. The anti-aliasing method prevents neural representation from learning aliased signals. }
   \label{fig:a_aliasing}
\end{figure}

\textbf{Anti-Aliasing is Efficient Enough}.
Compared to training, we use denser coordinates to get the ``interpolated'' result of an image, in other words, to achieve super-resolution. As shown in \cref{fig:a_aliasing}, introducing variational coordinates noticeably mitigates the negative effects caused by overly high-frequency inputs. As discussed before, MLP-based INR also suffers from the aliasing problem. With the anti-aliasing method, both ANR and MLP-based INR that are directly generated by a hyper-network can simulate a smooth reconstruction result.
\par
\textbf{Threshold boosts Localized Attention}.
During the attention alignment process, we utilize a threshold to clip those small attention weight values to zero. We compare the experiments on whether applying a localized attention layer instead of a normal attention layer which is discussed in \cref{sec:threshold_in_lal}. We find that a thresholded attention map that stops the irrelevant backpropagated gradient does help to accelerate the convergence process.
Subsequently, a local pattern of data will only guide the training of some representing vectors, enabling a better reconstruction.
\par
\begin{figure}[t]
  \centering
   \includegraphics[width=0.65\linewidth]{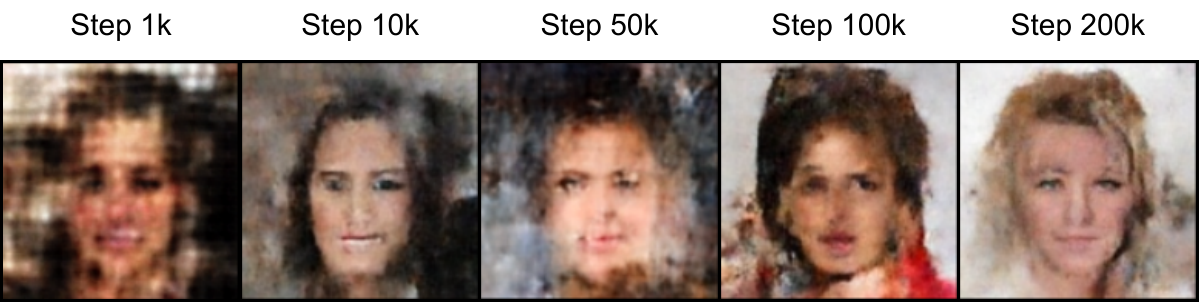}
   \caption{The diffusion sampling examples while the model is trained for different steps.}
   \label{fig:dif_gen_celeba}
\end{figure}

\textbf{Learning Downstream Tasks utilizing ANR as Prerequisite}.
To show that the ANR is a representation form that can be further used in downstream tasks, we provide an example experiment for diffusion generation tasks \cite{ho2020denoising}. As shown in \cref{fig:dif_gen_celeba}, we test diffusion training on the CelebA dataset and provide the diffusion sampling examples while the model is trained for different steps.
The main modification for this experiment is inserting a token containing time information into the R-Tokens. Through the modification, we get the modified R-Tokens containing both time and data information. Because those tokens are the intermediate outputs of the transformer, we can effortlessly denoise these tokens using the transformer. 
It's worth noting that the representation size could be one limitation of a generative model's capability. Owing to space constraints, we only use the instance-specific representation vectors (R-Tokens) of ANR as the representation form. We provide more details in \cref{sec:ds_task}.

\par
\par

\section{Conclusion}
\label{sec:conclusion}

\par
We propose \textbf{A}ttention-based Localized implicit \textbf{N}eural \textbf{R}epresentation (ANR), an innovative framework that significantly enhances the efficiency of neural representation, particularly in accommodating substantial and high-frequency data. 
Relying on feature tokens for data representation rather than the conventional weight and bias parameters of MLP, ANR marks a departure from traditional INRs. 
We introduce the L-softmax alignment, a solution to the pervasive issue of attention map generation in conventional softmax alignment. By constraining insignificant attention weights to zero, this mechanism enables a more focused and efficient training procedure. 
Moreover, the utilization of variational coordinates in addressing the aliasing problem within ANRs has demonstrated its efficacy in enhancing the robustness and accuracy of the model.
The extensive experiments not only validate the efficacy of ANRs but also underscore their superiority over conventional MLP-based INRs. 
The observed faster convergence rates and superior handling of diverse data modalities, such as 3D shapes, serve as compelling evidence of ANR's robustness and versatility.
\par
In essence, the contributions made in this study elucidate the potential of ANRs in revolutionizing neural representation learning, offering a more efficient and adaptable framework capable of addressing complex data patterns across multiple domains.
ANRs represent a promising avenue for future exploration and development.

\clearpage
\section*{Acknowledgements}
This work is supported by the National Key R\&D Program of China (Grant No. 2022ZD0160703), the National Natural Science Foundation of China (Grant No. 62202422), the Natural Science Foundation of Shandong Province (Grant No. ZR2021MH227), and Shanghai Artificial Intelligence Laboratory.

%
%

\bibliographystyle{splncs04}
\bibliography{main}

\appendix
\input{X_suppl}

\end{document}

%% file: X_suppl.tex
\clearpage
\setcounter{page}{1}
{
   \newpage
        \centering
        \Large
        \vspace{0.5em}Supplementary Material \\
        \vspace{1.0em}
}
\section{The Workflow of the Hyper-network}
\label{sec:trc}

\begin{figure}[!t]
  \centering
   \includegraphics[width=0.6\linewidth]{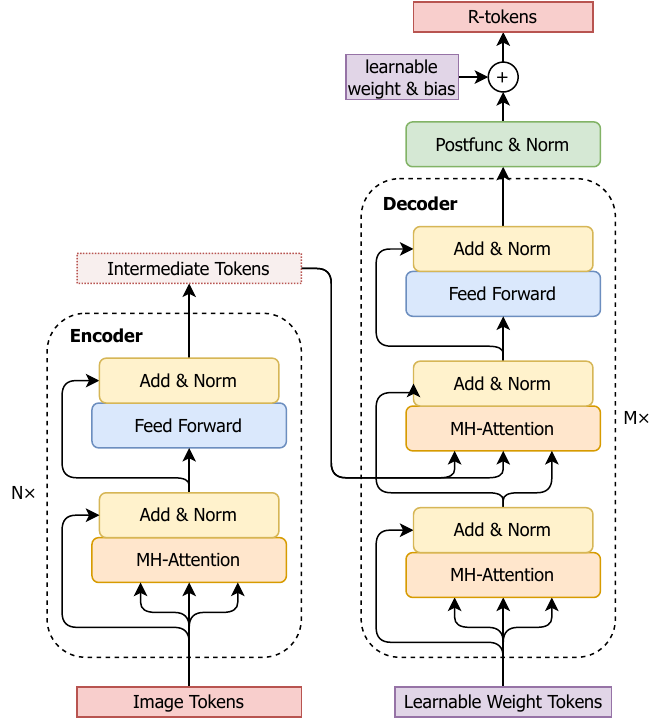}
   \caption{The architecture of hyper-network that is used for R-tokens generating.}
   \label{fig:hyper_arc}
\end{figure}

We design a Transformer-like hyper-network for predicting R-Tokens, which consists of a data instance encoder and a neural representation decoder. The encoder processes data information and generates feature tokens, while the decoder converts the processed feature tokens into R-Tokens. The encoder directly utilizes a standard Transformer encoder, while the decoder differs from a standard Transformer decoder in omitting the masked self-attention layer. We illustrate the computation process in \cref{fig:hyper_arc}.
\par
Different from Chen~\etal \cite{chen2022transformers} that uses only one Transformer encoder to generate INRs, our model can independently decide the depth of the encoder and the decoder according to data size and the design size of R-Tokens, respectively. Furthermore, in the long term, the intermediate variables of the encoder's output can be considered structured data features so that we can integrate them into other networks as input for downstream tasks.


\textbf{Hyper-parameters}. The default configuration of hyper-parameters in the experiments includes a 6-head attention mechanism, where each head possesses a feature dimension of 64. Additionally, the hidden dimension of each feed-forward layer is set to 3072. The depth of both the encoder and decoder is by default set to 6 for all experiments.


\subsection{Discussion of the Hyper-network Architecture}
We compare the speed of convergence between our designed Transformer-like hyper-network architecture and the encoder-only hyper-network architecture used by Chen~\etal~\cite{chen2022transformers} on the CelebA dataset. 
\par
\textbf{Analysis of Runtime Memory Efficiency}.
Denoting the width of the vector of the tokenized data by $L_d$ and the width of the target representation feature vector by $L_r$, we ascertain that the size of the attention map of each layer of the encoder-only hyper-network is $(L_d+L_r)\times(L_d+L_r)$. 
In contrast, within a transformer-like hyper-network, the attention map with the size of $L_d \times L_r$, $L_d\times L_d$ and $L_r\times L_r$ are designed for cross-attention, encoder's self-attention, and encoder's self-attention, respectively.
Consequently, the transformer-like hyper-network's attention map is comparatively smaller, indicating running time memory efficiency, as shown in \cref{tab:hyper_depth}. Note that the parameter sizes of ``Transformer6+6" and ``EncoderOnly12" are similar.
\par
\begin{figure}[!t]
  \centering
   \includegraphics[width=0.5\linewidth]{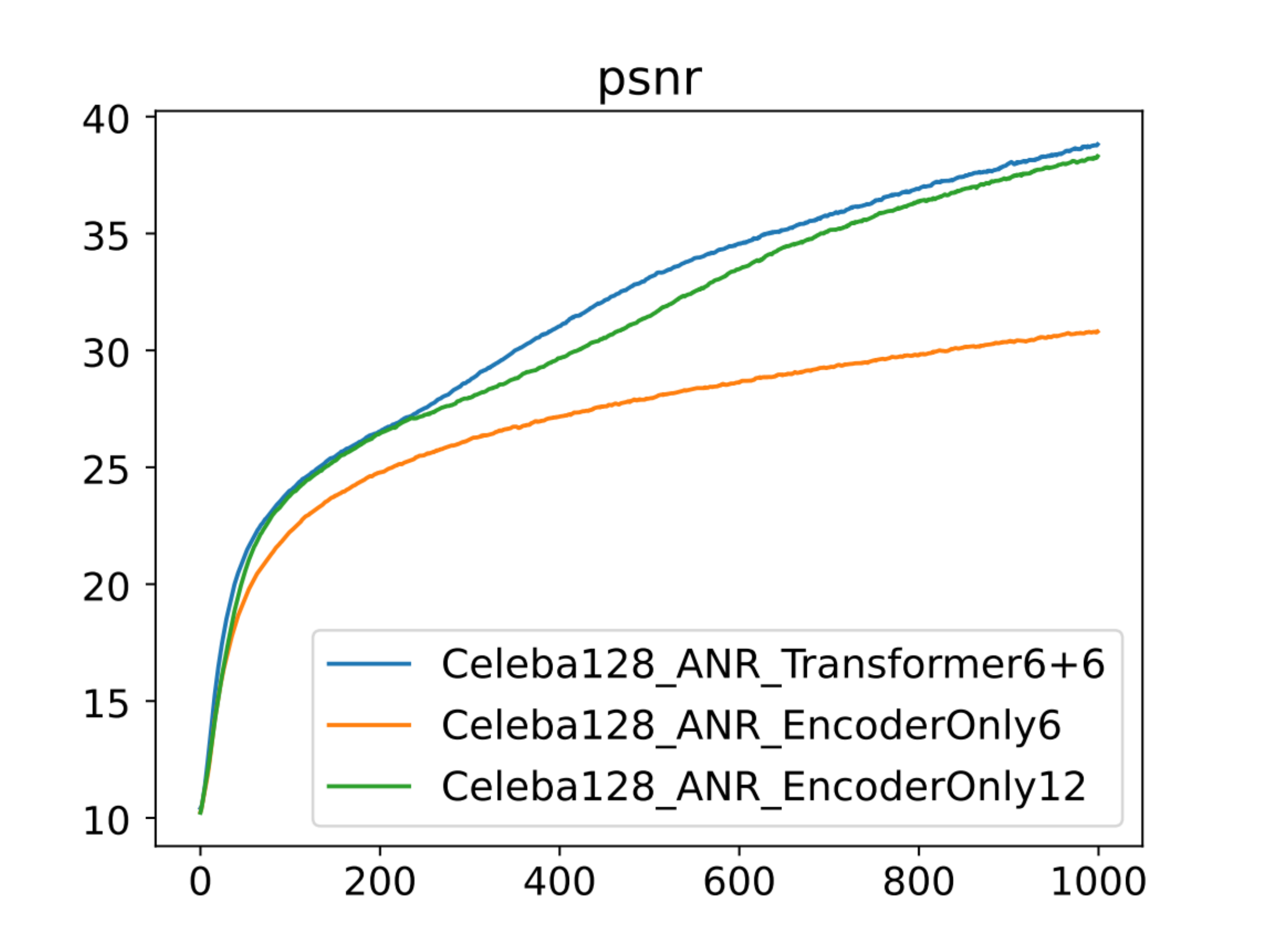}
   \caption{The PSNR curves of the ANR training process on the CelebA dataset.}
   \label{fig:comp_enconly}
\end{figure}

 
\textbf{Comparison Results of Runtime Efficiency}.
Except for the network depth, all the other hyper-parameters across the two compared architectures remain consistent. Specifically, the experiment setting of the corresponding architecture is presented in \cref{tab:hyper_depth}. 
We train these models for 100k steps. The experiment shows that the PSNR value ascends faster with a Transformer-like hyper-network, implying our design offers better performance, as shown in \cref{fig:comp_enconly}. 
\par

\begin{table*}[!t]
    \centering
    \caption{The comparison details of three hyper-networks.}
    \begin{tabular}{ccccl}
    \midrule
          Architecture&\makecell{Encoder\\Depth}&\makecell{Decoder\\Depth}  & \makecell{The Size of \\Attention Map} & \makecell{Memory\\Consumption}\\
    \midrule[2pt]
          transformer-like&6& 6  & $6\times(L_d^2+L_r^2+L_d\times L_r)$ &11409MiB\\
    \midrule
 encoder-only& 6&None  & $6\times(L_d^2+L_r^2+2\times L_d\times L_r)$&10899MiB\\
    \midrule
 encoder-only& 12&None  & $12\times(L_d^2+L_r^2+2\times L_d\times L_r)$&11709MiB\\
    \midrule
 \end{tabular}
    \label{tab:hyper_depth}
\end{table*}

\par

\section{Additional Implementation Details}
\label{sec:impl_details}
\textbf{Experimental Settings.}
The experiments are conducted using a workstation equipped with an NVIDIA RTX A6000 GPU. The software environment is configured with Python 3.8.18, CUDA 11.7, and PyTorch 1.10.0 as the primary deep learning framework. 
\par

\textbf{Model Architecture and Hyperparameters.}
We adopt a Transformer-like hyper-network to predict the feature vectors that contain the data information and are used to represent it. The detailed structure and configurations of the hyper-network are presented in \cref{sec:trc}. ANR receives the feature vectors (R-Tokens) and uses transform matrices to convert them to attention 
$K$, and $V$ and fuses them with coordinate features in the forward process. 

The $v_{dev}$ in \cref{eqa:coord} is set to $1 / (2*\texttt{max}(h,w))$ and the $\alpha$ is set to $3 / (20*\texttt{max}(h,w))$ where $h$ and $w$ are the data instance training resolution. The standard deviation $\sigma$ is set to 1.

\par

\section{Detail discussion of the LAL Forward Process.}
\label{sec:lal_disc_detail}
Firstly, the LAL layer first employs the dot product to integrate information from the data’s representation tokens with the query signal. In the dot product operation, the Key matrix can be viewed as a transposed weight matrix of a linear layer and multiplied with the Query matrix, thus adhering to \cref{prop:prop1}. 
\par
Subsequently, the softmax function is applied. We split the softmax computation into two parts: exponential function and normalization, where the exponential function portion follows \cref{prop:prop2}. For convenience, we denote the signal spectrum of the exponential function output as $\Omega_r$.
\par
Afterward, the signals undergo a sequence of operations: normalization, filtering, and re-normalization. During normalization operations, the wave spectrum is uniformly scaled; in the filtering operation, parts of neurons' values are directly filtered to zero.
These procedures do not alter the internal composition of the signals within one neuron but affect the overall signal amplitude.
\par
Therefore, we can still consider the neurons to be composed of sets of waves $\Psi(x, \widetilde{\Omega_r} )$, with the spectrum restructured to $\widetilde{\Omega_r}$. 
These signals are then input into the subsequent MLP, ultimately yielding the final output.
\par
Consequently, the ANR forwarding process should have the form:
\begin{equation}
\label{eqa:anr_series}
\begin{aligned} 
f_\theta(c,D)&=\texttt{ANR}(\gamma(c),D) = \texttt{MLP}\circ \texttt{LAL}(W_q \Psi(c, \Omega ) ,W_kD,W_vD) \\
&= \texttt{MLP}(\Psi(c, \widetilde{\Omega_r} )) = \Psi(c, \widetilde{\Omega_r}^{\prime}).
\end{aligned} 
\end{equation}
\par
From \cref{eqa:anr_series}, we can infer that ANRs can considered as INRs with the filtered basic spectrum $\widetilde{\Omega_r}$, because INR can be thought of as a structured dictionary \cite{yuce2022structured} or Fourier series \cite{benbarka2022seeing} with the form:
\begin{equation}
\label{eqa:inr_series}
f_{\theta}(c)=\texttt{INR}(\gamma(c))=\texttt{MLP}(\Psi(c, \Omega ))\\
=\Psi(c, \Omega^{\prime} ).
\end{equation}
\par
In summary, the LAL layer first fuses the information of query signal waves with the data's representation tokens, then filters and reweights them. The LAL selects the most relevant wave signals for the current query. With these signals, ANRs can reconstruct target signals better.

\section{Experiments on Low-Frequency Positional Embedding}
\label{sec:lowf_exp}
Due to the requirement that neural representations should be able to encode continuous functions, we conduct experiments to demonstrate the feasibility of using ANR to represent a continuous function.
Given that the INR Function employs a coordinate-based mapping, we can verify the continuity of the INR Function by examining the coherence of the interpolation between the originally sampled coordinates
In other words, we consider ANR continuous if its output remains continuous and reasonable when our sampled points extend beyond the training points.
\par
Additionally, we note that all INR Functions, including MLP-based INR, may suffer from aliasing due to high frequencies in the positional embedding (P.E.) of coordinates, as depicted in \cref{fig:a_aliasing}. 
\par
\subsection{Whether ANR is a Continuous Function}
\begin{figure}[!t]
  \centering
   \includegraphics[width=0.65\linewidth]{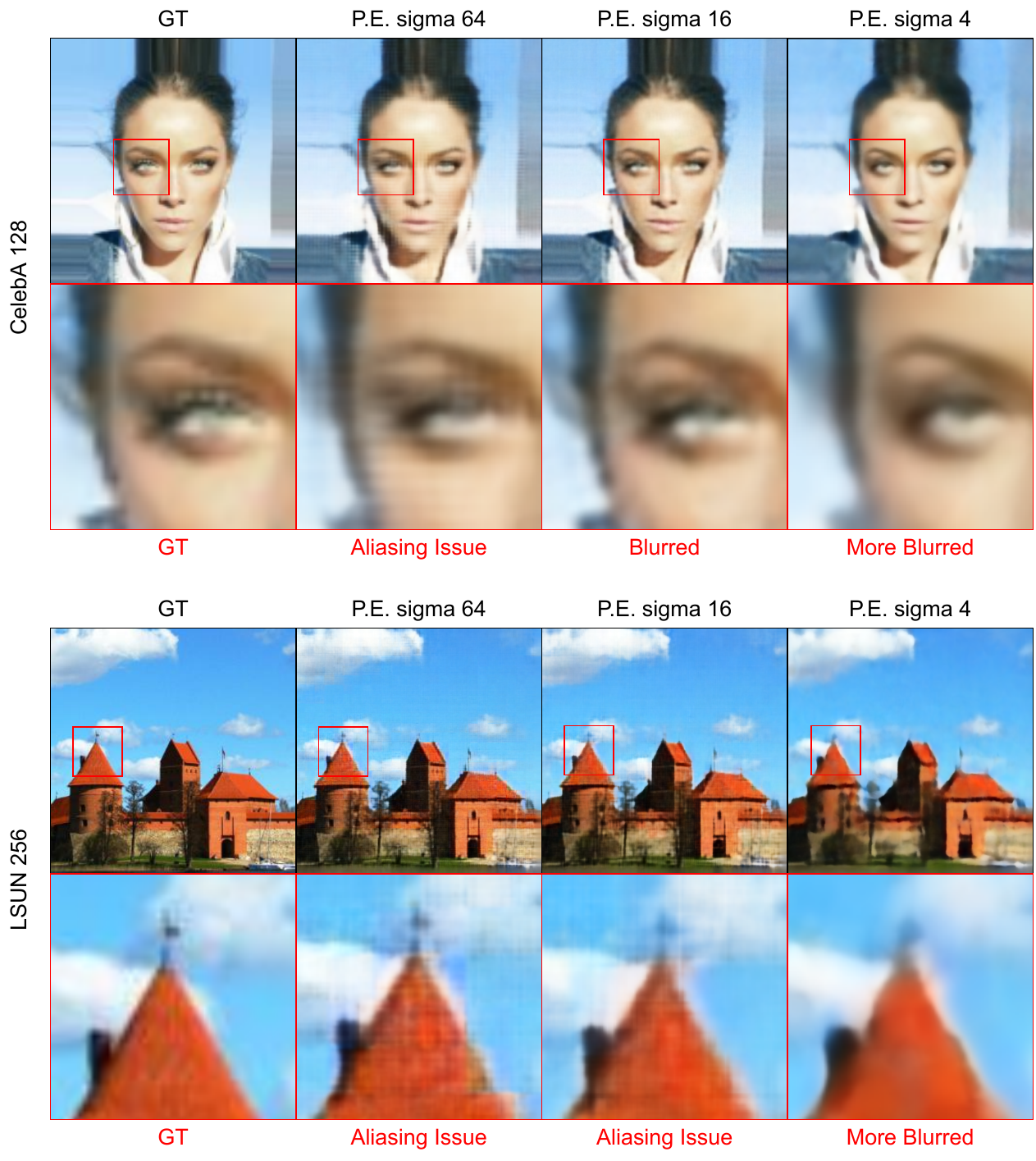}
   \caption{The visualization depicts experiments on coordinate interpolation with a low frequency of positional encoding (P.E.). When the P.E. frequency is kept sufficiently low to avoid aliasing effects, the outputs from the ANR remain continuous after the interpolation of coordinates. The term `P.E. sigma' in the figure denotes the frequency of coordinates. Images bordered in red highlight the details of the original images. }
   \label{fig:lowf_coordpe}
\end{figure}

When testing whether ANR is a continuous function, we set the frequency of P.E. to be sufficiently low to alleviate aliasing. We conducted experiments on the LSUN dataset and the CelebA dataset separately. Because the capacity of INR is limited with low P.E. frequency, we set the number of training steps to 40k, which we found to be adequate for convergence.
As illustrated in \cref{fig:lowf_coordpe}, our ANR outputs remain continuous after coordinate interpolation. This demonstrates the ability of ANR to encode continuous functions. Note that we train the model without anti-aliasing methods.

\section{Periodic Activation Function Performance}
\label{sec:periodic_act}
The use of periodic activation necessitates initializing all weights in MLP layers from a uniform distribution, $W \sim U(-\sqrt{fan_{in}},\sqrt{fan_{in}})$ \cite{sitzmann2020implicit}. However, it's challenging to constrain the distribution of MLP weights predicted by the hyper-network. One solution is introducing modulation weights \cite{mehta2021modulated}, but it requires additional effort for training other modules. 
\par
We have conducted experiments with periodic activation, which showed unstable convergence results due to the pretty high non-linearity of periodic activation. The comparison is provided in \cref{tab:periodic_exp}.

\begin{table*}[t]
  \centering
  \caption{Performance of different activation functions in used neural representation.}
  \begin{tabular}{ccccc}
    \toprule
       Dataset& Method&  Activation&MLP Depth& PSNR$\uparrow$ \\
    \midrule[2pt]
    \multirow{2}*{Celeba}& ANR& ReLU&5& \textbf{47.25}\\
      & ANR& Sine&5& 23.45\\
    \midrule
    \multirow{2}*{Celeba} & INR& ReLU&12& 27.54\\ 
      &  INR& Sine&12& 25.63\\
    \midrule
  \end{tabular}
  \label{tab:periodic_exp}
\end{table*}

\section{Solving Aliasing Problem in 1D Signal Fitting.}
\label{sec:1d_aliasing}
To better understand how a variational coordinate sampling strategy works in real problems, we provide a 1D signal-fitting example.
\par
First, we will introduce the experimental setup. For simplicity and representativeness, we choose a Multi-Layer Perceptron (MLP) as the network model to be trained, and we will use Adam as the optimizer. We set our original input as a vector uniformly distributed between $(-1, 1)$ with a spacing of $2/N$, where $N$ is the number of sampling points. Additionally, the output is designed as a combination of trigonometric function signals. We ensure that the frequency of the trigonometric functions is not too high to avoid under-sampling issues. In the experiment, we set the number of sampling points to 100 and the target signal is the combination of 10 randomly chosen frequency signals.
\par
Next, we train our network model using Mean Squared Error (MSE) Loss. We will train two identical MLPs independently, each with 5 hidden layers and a dimension of 256. One network will be trained for 400 steps with original input. For the other network, we introduce coordinate Gaussian Shifts as described in the main text to expand the sampling space of the network’s input. Specifically, after sampling Gaussian noise from a standard distribution, we multiply it by a very small scaling factor $s=1/(5N)$ and add it to the original input. This network will also be trained for 400 steps.
\par
\begin{figure}[!t]
  \centering
   \includegraphics[width=0.6\linewidth]{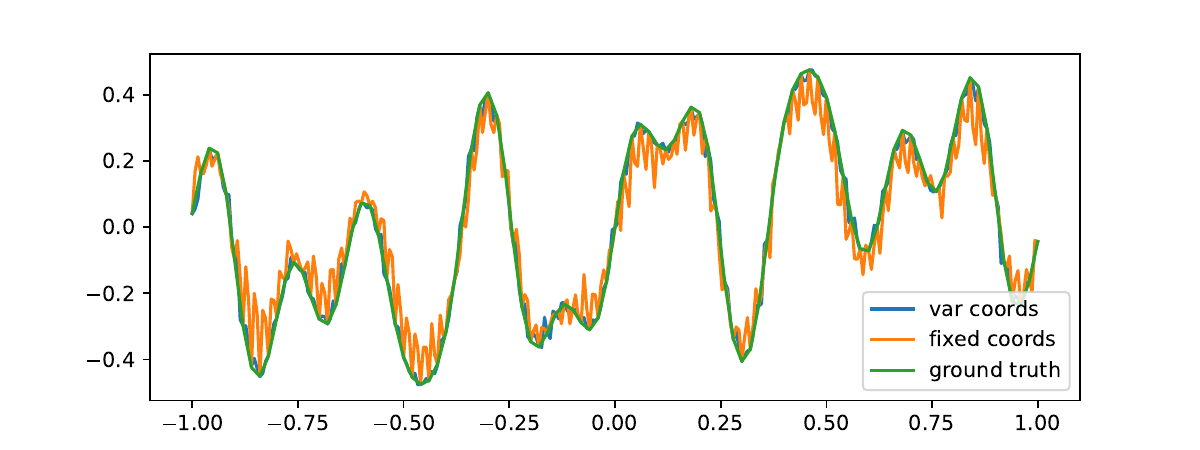}
   \caption{The super-resolution result of two networks on the 1D signal fitting problem. The model trained with fixed coordinates learned fake signal frequencies and the interpolation between coordinates is meaningless. While the model trained with resampled coordinates significantly suppresses the abnormal frequencies.}
   \label{fig:1dfit}
\end{figure}

As shown in \cref{fig:1dfit}, we find that the network using the variational coordinate sampling strategy is less likely to learn fake high-frequency signals when predicting super-resolution signals. It indicates that the method significantly suppresses the aliasing problem, as we proposed.

\section{Inference Examples on the ERA5 Dataset}
\begin{figure*}[ht]
  \centering

  \includegraphics[width=0.75\linewidth]{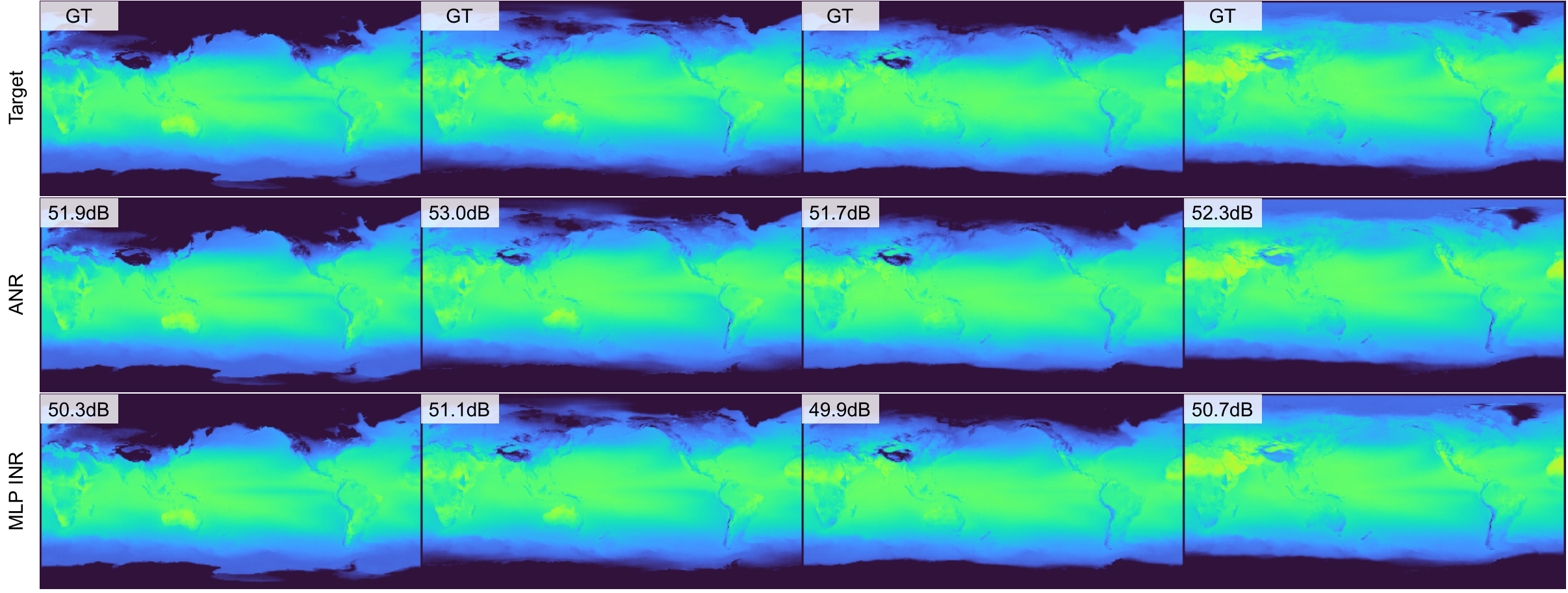}
  \caption{The inference examples on the ERA5 dataset. The target images contain relatively invariant features, such as edges. }
  \label{fig:exp_on_era5}
\end{figure*}

The inference examples on the ERA5 dataset are shown in \cref{fig:exp_on_era5}.  The neural representation function is designed for predicting the surface temperature of a specific location on Earth, while the input coordinates are in latitude and longitude. The Earth's temperature field greatly depends on its physical structures, resulting in similar border patterns in the target image. Experimental findings demonstrate that the ANR and MLP INR models can efficiently represent these relatively invariant characteristics.
\label{sec:inf_example_era5}

\section{Threshold Boosts the Localized Attention}
\label{sec:threshold_in_lal}
We propose the localized attention layer (LAL) and discuss it in \cref{sec:lal_arc}. An LAL replaces the origin attention alignment process (softmax alignment) with L-softmax alignment. L-softmax adds a threshold $m$ as attention map activation so that the attention weights smaller than the threshold are clipped to zero, and the weights are normalized again. In our experimental investigations, we determine that setting the parameter m to 0.0015 yields good performance. 
\par
We compare the reconstruction performance of ANR on Celeba and LSUN datasets, as shown in \cref{tab:ANR_threshold_eff}. We set the depth of the MLP representation converter large enough to fit real application situations. Experiment results show that the ANR convergence speed of those with attention threshold is faster than those without. 
\begin{table*}[t]
  \centering
  \caption{The comparison result of whether the attention layer is localized or not.}
  \begin{tabular}{cccccc}
    \toprule
       Dataset&MLP Depth& \makecell{Loss$\downarrow$\\ (w/ threshold)}& \makecell{Loss$\downarrow$\\ (w/o threshold)}& \makecell{PSNR$\uparrow$\\ (w/ threshold)}& \makecell{PSNR$\uparrow$\\ (w/o threshold)}\\
    \midrule[2pt]
    \multirow{2}*{Celeba}&3&\textbf{1.1e-4}& 1.6e-4& \textbf{40.25}&38.65\\
      &5&\textbf{2.2e-5}& 4.5e-5& \textbf{47.25}&44.21\\
    \midrule
       \multirow{2}*{LSUN}&5&\textbf{2.2e-3}&3.1e-3& \textbf{27.54}&26.05\\ 
      & 7& \textbf{1.9e-3}& 2.4e-3& \textbf{28.30}&27.16\\
    \midrule
  \end{tabular}
  \label{tab:ANR_threshold_eff}
\end{table*}

\section{Learning Downstream Tasks utilizing ANR as Prerequisite}
\label{sec:ds_task}

\begin{figure*}[t]
  \centering

  \includegraphics[width=0.8\linewidth]{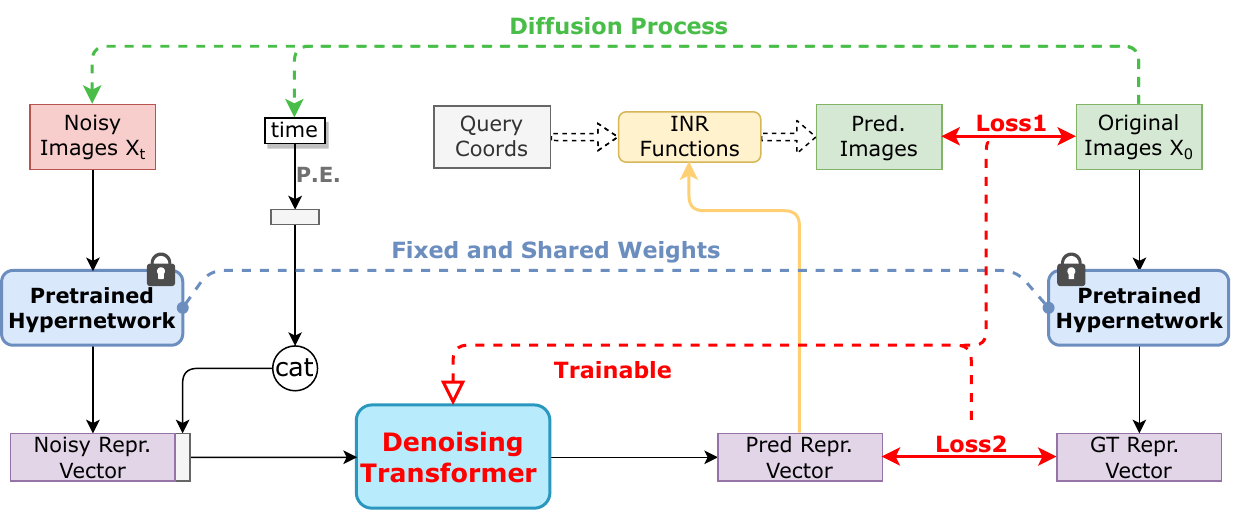}
  \caption{The diffusion training framework utilizes ANR as the data representation form. The pre-trained hypernetwork is fixed, and some additional Transformer blocks serve as the trainable denoiser. The loss is calculated from both the predicted images and the original images (Loss1), as well as from the predicted representation vectors and the original representation vectors (Loss2). 
  Note that before calculating the Loss2, the representation vectors incorporate L2 normalization on their last dimension. Subsequently, we rescale this normalized loss by multiplying it by the size of the vectors's last dimension. Finally, we backpropagate the average loss computed between Loss1 and Loss2. This approach ensures a balanced treatment of the two losses.}
  \label{fig:gen_fw}
\end{figure*}

We provide details regarding utilizing ANR as a prerequisite for downstream tasks. We present an experiment on diffusion generation tasks, utilizing the CelebA dataset. The experiment involves modifying R-Tokens to incorporate time information, resulting in representations containing both temporal and data features. These modified R-Tokens, being intermediate outputs of the transformer, facilitate effortless denoising within the transformer framework.
\par
\begin{figure*}[t]
  \centering

  \includegraphics[width=0.975\linewidth]{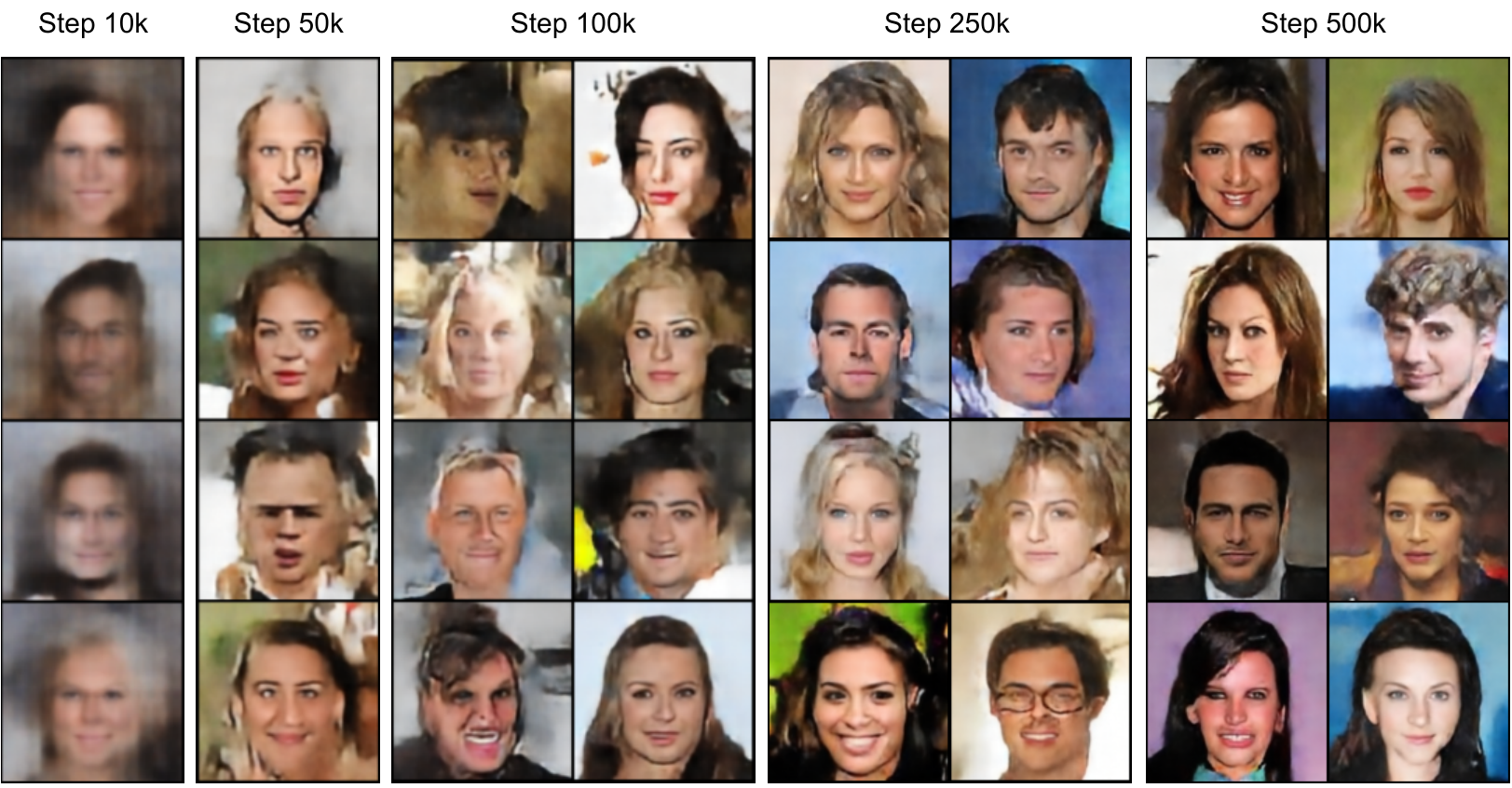}
  \caption{More diffusion sampling examples while the model is trained for different steps. With the training procedure, the sample quality is improving and the sampled images could contain more details and fewer abnormal structures.}
  \label{fig:celeba_gen_result_example}
\end{figure*}

As illustrated in \cref{fig:gen_fw}, we train the model under the diffusion training framework. The denoising Transformer consists of several Transformer Blocks, each containing a self-attention layer and a feed-forward layer. Because we want to demonstrate that the representation form can be directly used for downstream tasks, we do not apply any structural modifications to the denoising Transformer such as more skip connections. Moreover, we do not assume additional spatial relationships across the tokens, even though they might enhance the generation performance. 
In the experiments, we set the batch size higher than that used for the reconstruction task, at 48. The number of denoising Transformer Blocks is 8, and the feature dimension of each attention head is set to 96. All other hyper-parameters remain consistent with those of the reconstruction task, as detailed in \cref{sec:trc}. We find that the diffusion results are still impressive, as shown in \cref{fig:celeba_gen_result_example}.

\section{Extensive Results on Diverse Datasets}
\label{sec:more_res}
\begin{figure*}[t]
  \centering

  \includegraphics[width=0.85\linewidth]{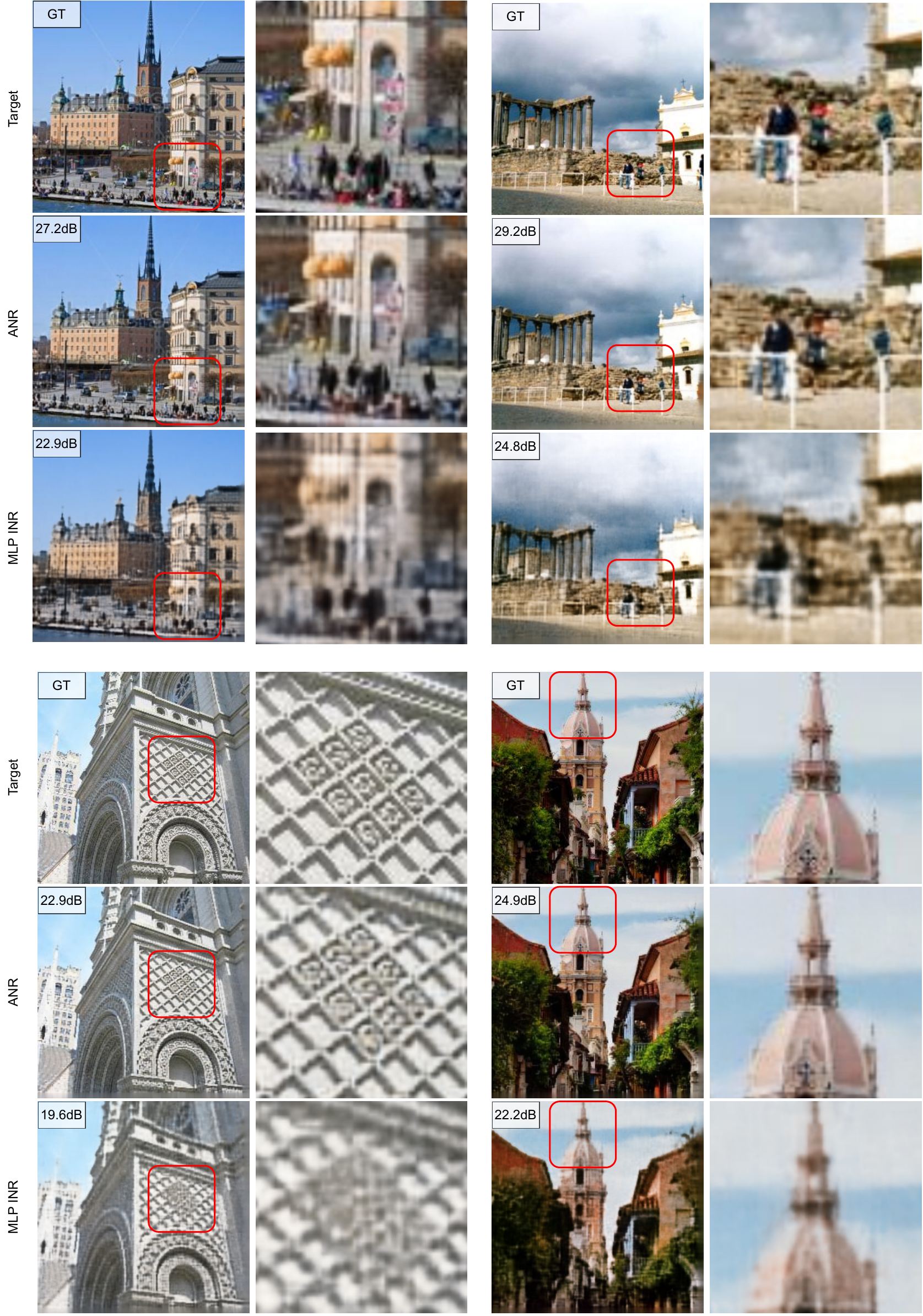}
  \caption{More inference results on LSUN. With ANR, the details and colors of the building have been better restored. ( The images on the right side show the enlarged details of the original image. )}
  \label{fig:more_res_lsun}
\end{figure*}

\begin{figure*}[ht]
  \centering

  \includegraphics[width=\linewidth]{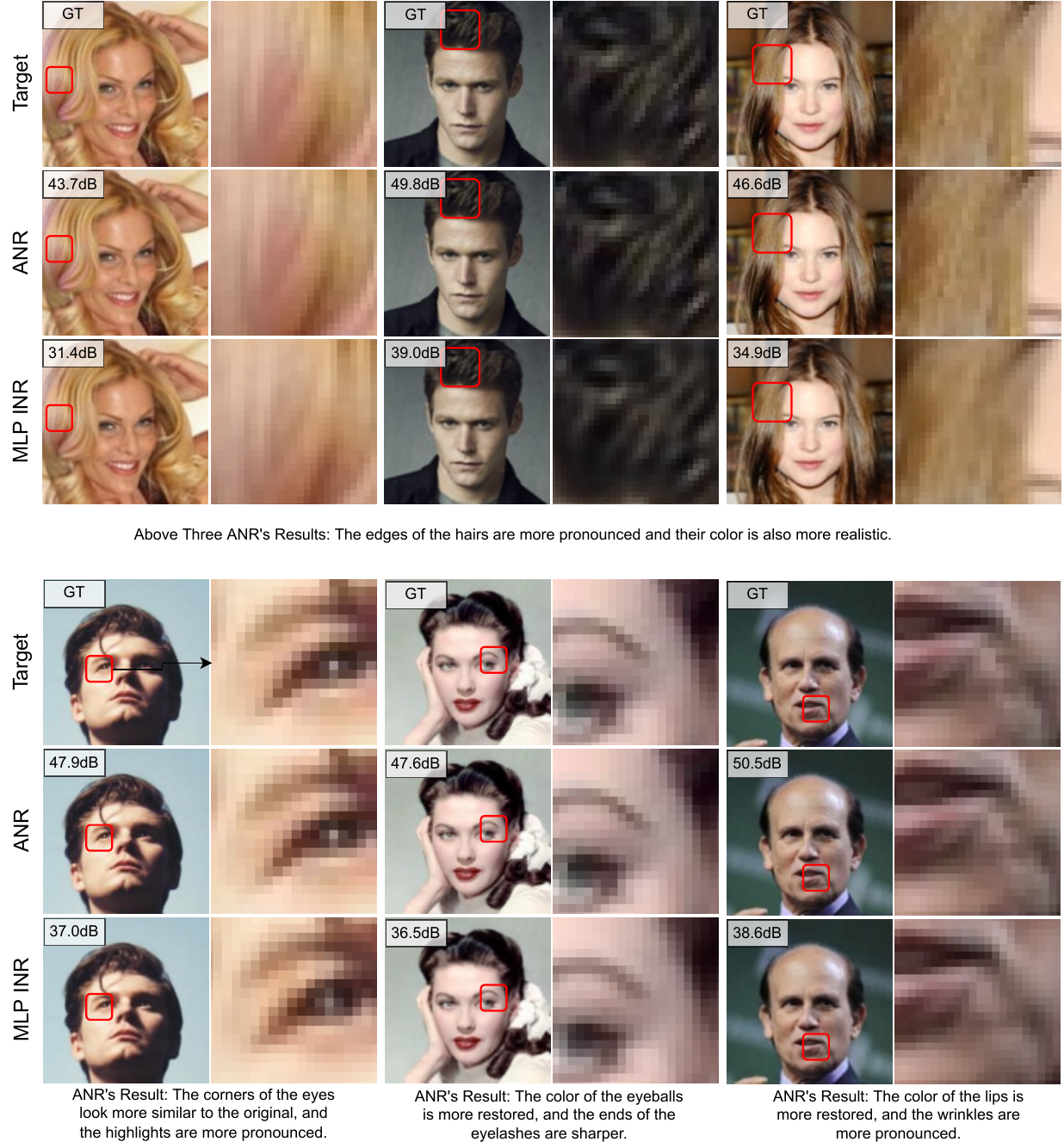}
  \caption{More inference results on CelebA. The results show that ANR can reconstruct image details better, especially the places where there are lots of lines and corners (for example, the person's hairs and eyes) implying the existence of high-frequency signals. ( The images on the right side show the enlarged details of the original image. ) }
  \label{fig:more_res_celeba}
\end{figure*}

\begin{figure*}[t]
  \centering

  \includegraphics[width=\linewidth]{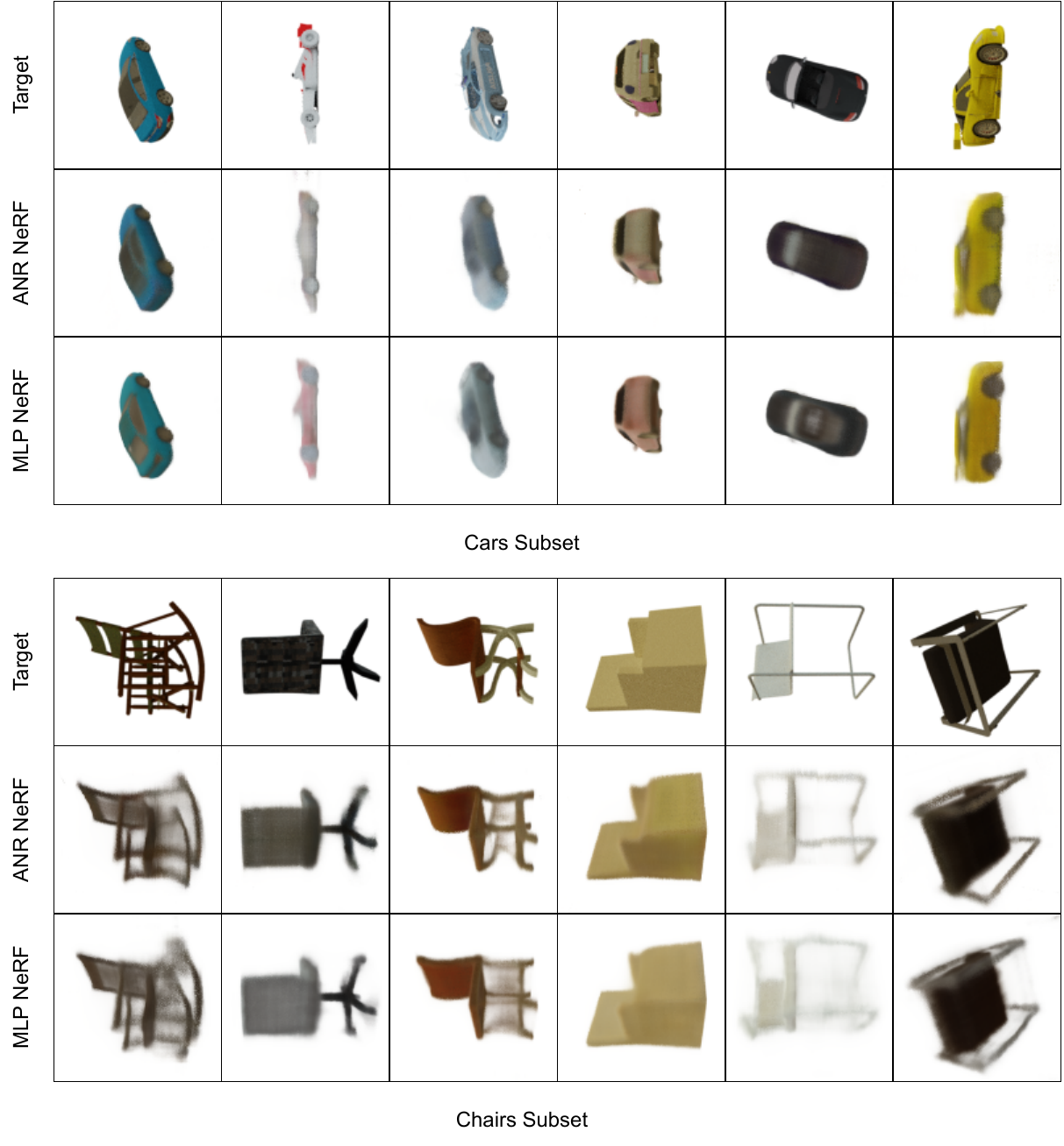}
  \caption{More inference results on Learnit ShapeNet. We found that the view synthesis results using ANR have more accurate shape boundaries, and the MLP-based NeRF tends to use the color of parts of the object to predict the overall color.}
  \label{fig:more_res_shapenet}
\end{figure*}

In this supplementary section, we present additional results and analyses for the performance evaluation of our proposed model across diverse datasets: CelebA, LSUN, and Learnit Shapenet. Due to spatial constraints in the main body of the paper, we provide an extended array of qualitative visualizations showcasing the efficacy of our model in image reconstruction and view synthesis tasks. The experiments conducted demonstrate notable improvements achieved by our model compared to existing methods, particularly highlighting its superior performance in image reconstruction tasks across various datasets, as shown in \cref{fig:more_res_celeba} and \cref{fig:more_res_lsun}. Furthermore, our model's efficiency in view synthesis tasks is elucidated through comprehensive visual representations, as shown in \cref{fig:more_res_shapenet}, illustrating its capacity to generate high-quality synthesized views with enhanced accuracy and realism.


%% file: main.bbl
\begin{thebibliography}{10}
\providecommand{\url}[1]{\texttt{#1}}
\providecommand{\urlprefix}{URL }
\providecommand{\doi}[1]{https://doi.org/#1}

\bibitem{anciukevivcius2023renderdiffusion}
Anciukevi{\v{c}}ius, T., Xu, Z., Fisher, M., Henderson, P., Bilen, H., Mitra, N.J., Guerrero, P.: Renderdiffusion: Image diffusion for 3d reconstruction, inpainting and generation. In: Proceedings of the IEEE/CVF Conference on Computer Vision and Pattern Recognition. pp. 12608--12618 (2023)

\bibitem{bauer2023spatial}
Bauer, M., Dupont, E., Brock, A., Rosenbaum, D., Schwarz, J.R., Kim, H.: Spatial functa: Scaling functa to imagenet classification and generation. arXiv preprint arXiv:2302.03130  (2023)

\bibitem{benbarka2022seeing}
Benbarka, N., H{\"o}fer, T., Zell, A., et~al.: Seeing implicit neural representations as fourier series. In: Proceedings of the IEEE/CVF Winter Conference on Applications of Computer Vision. pp. 2041--2050 (2022)

\bibitem{brauwers2021general}
Brauwers, G., Frasincar, F.: A general survey on attention mechanisms in deep learning. IEEE Transactions on Knowledge and Data Engineering  (2021)

\bibitem{chan2021pi}
Chan, E.R., Monteiro, M., Kellnhofer, P., Wu, J., Wetzstein, G.: pi-gan: Periodic implicit generative adversarial networks for 3d-aware image synthesis. In: Proceedings of the IEEE/CVF conference on computer vision and pattern recognition. pp. 5799--5809 (2021)

\bibitem{chang2015shapenet}
Chang, A.X., Funkhouser, T., Guibas, L., Hanrahan, P., Huang, Q., Li, Z., Savarese, S., Savva, M., Song, S., Su, H., et~al.: Shapenet: An information-rich 3d model repository. arXiv preprint arXiv:1512.03012  (2015)

\bibitem{chen2022transformers}
Chen, Y., Wang, X.: Transformers as meta-learners for implicit neural representations. In: European Conference on Computer Vision. pp. 170--187. Springer (2022)

\bibitem{chen2021neural}
Chen, Y., Fernando, B., Bilen, H., Mensink, T., Gavves, E.: Neural feature matching in implicit 3d representations  (2021)

\bibitem{chen2022videoinr}
Chen, Z., Chen, Y., Liu, J., Xu, X., Goel, V., Wang, Z., Shi, H., Wang, X.: Videoinr: Learning video implicit neural representation for continuous space-time super-resolution. In: Proceedings of the IEEE/CVF Conference on Computer Vision and Pattern Recognition. pp. 2047--2057 (2022)

\bibitem{dupont2022data}
Dupont, E., Kim, H., Eslami, S.A., Rezende, D.J., Rosenbaum, D.: From data to functa: Your data point is a function and you can treat it like one. In: International Conference on Machine Learning. pp. 5694--5725. PMLR (2022)

\bibitem{hersbach2019era5}
Hersbach, H., Bell, B., Berrisford, P., Biavati, G., Hor{\'a}nyi, A., Mu{\~n}oz~Sabater, J., Nicolas, J., Peubey, C., Radu, R., Rozum, I., et~al.: Era5 monthly averaged data on single levels from 1979 to present. Copernicus Climate Change Service (C3S) Climate Data Store (CDS)  \textbf{10},  252--266 (2019), \url{https://cds.climate.copernicus.eu/cdsapp\#!/dataset/reanalysis-era5-single-levels-monthly-means?tab=form}(Accessed 22-09-2023)

\bibitem{ho2020denoising}
Ho, J., Jain, A., Abbeel, P.: Denoising diffusion probabilistic models. In: Proceedings of the 34th International Conference on Neural Information Processing Systems. pp. 6840--6851 (2020)

\bibitem{jun2023shap}
Jun, H., Nichol, A.: Shap-e: Generating conditional 3d implicit functions. arXiv preprint arXiv:2305.02463  (2023)

\bibitem{kim2023generalizable}
Kim, C., Lee, D., Kim, S., Cho, M., Han, W.S.: Generalizable implicit neural representations via instance pattern composers. In: Proceedings of the IEEE/CVF Conference on Computer Vision and Pattern Recognition. pp. 11808--11817 (2023)

\bibitem{kim2021lipschitz}
Kim, H., Papamakarios, G., Mnih, A.: The lipschitz constant of self-attention. In: International Conference on Machine Learning. pp. 5562--5571. PMLR (2021)

\bibitem{kingma2013auto}
Kingma, D.P., Welling, M.: Auto-encoding variational bayes. arXiv preprint arXiv:1312.6114  (2013)

\bibitem{landgraf2022pins}
Landgraf, Z., Hornung, A.S., Cabral, R.S.: Pins: Progressive implicit networks for multi-scale neural representations. In: International Conference on Machine Learning. pp. 11969--11984. PMLR (2022)

\bibitem{larochelle2010learning}
Larochelle, H., Hinton, G.E.: Learning to combine foveal glimpses with a third-order boltzmann machine. Advances in neural information processing systems  \textbf{23} (2010)

\bibitem{lee2024locality}
Lee, D., Kim, C., Cho, M., HAN, W.S.: Locality-aware generalizable implicit neural representation. Advances in Neural Information Processing Systems  \textbf{36} (2024)

\bibitem{lin2021barf}
Lin, C.H., Ma, W.C., Torralba, A., Lucey, S.: Barf: Bundle-adjusting neural radiance fields. In: Proceedings of the IEEE/CVF International Conference on Computer Vision. pp. 5741--5751 (2021)

\bibitem{liu2023partition}
Liu, K., Liu, F., Wang, H., Ma, N., Bu, J., Han, B.: Partition speeds up learning implicit neural representations based on exponential-increase hypothesis. In: Proceedings of the IEEE/CVF International Conference on Computer Vision. pp. 5474--5483 (2023)

\bibitem{liu2023implicit}
Liu, K., Ma, N., Wang, Z., Gu, J., Bu, J., Wang, H.: Implicit neural distance optimization for mesh neural subdivision. In: 2023 IEEE International Conference on Multimedia and Expo (ICME). pp. 2039--2044. IEEE (2023)

\bibitem{liu2015faceattributes}
Liu, Z., Luo, P., Wang, X., Tang, X.: Deep learning face attributes in the wild. In: Proceedings of International Conference on Computer Vision (ICCV) (December 2015)

\bibitem{lu2016hierarchical}
Lu, J., Yang, J., Batra, D., Parikh, D.: Hierarchical question-image co-attention for visual question answering. Advances in neural information processing systems  \textbf{29} (2016)

\bibitem{luong2015effective}
Luong, M.T., Pham, H., Manning, C.D.: Effective approaches to attention-based neural machine translation. arXiv preprint arXiv:1508.04025  (2015)

\bibitem{ma2017interactive}
Ma, D., Li, S., Zhang, X., Wang, H.: Interactive attention networks for aspect-level sentiment classification. In: Proceedings of the 26th International Joint Conference on Artificial Intelligence. pp. 4068--4074 (2017)

\bibitem{martel2021acorn}
Martel, J.N., Lindell, D.B., Lin, C.Z., Chan, E.R., Monteiro, M., Wetzstein, G.: Acorn: Adaptive coordinate networks for neural scene representation. arXiv preprint arXiv:2105.02788  (2021)

\bibitem{mehta2021modulated}
Mehta: Modulated periodic activations for generalizable local functional representations. In: ICCV (2021)

\bibitem{mildenhall2021nerf}
Mildenhall, B., Srinivasan, P.P., Tancik, M., Barron, J.T., Ramamoorthi, R., Ng, R.: Nerf: Representing scenes as neural radiance fields for view synthesis. Communications of the ACM  \textbf{65}(1),  99--106 (2021)

\bibitem{mnih2014recurrent}
Mnih, V., Heess, N., Graves, A., et~al.: Recurrent models of visual attention. Advances in neural information processing systems  \textbf{27} (2014)

\bibitem{rahaman2019spectral}
Rahaman, N., Baratin, A., Arpit, D., Draxler, F., Lin, M., Hamprecht, F., Bengio, Y., Courville, A.: On the spectral bias of neural networks. In: International conference on machine learning. pp. 5301--5310. PMLR (2019)

\bibitem{rebain2022attention}
Rebain, D., Matthews, M.J., Yi, K.M., Sharma, G., Lagun, D., Tagliasacchi, A.: Attention beats concatenation for conditioning neural fields. Transactions on Machine Learning Research  (2022)

\bibitem{reiser2021kilonerf}
Reiser, C., Peng, S., Liao, Y., Geiger, A.: Kilonerf: Speeding up neural radiance fields with thousands of tiny mlps. In: Proceedings of the IEEE/CVF International Conference on Computer Vision. pp. 14335--14345 (2021)

\bibitem{shen2018disan}
Shen, T., Zhou, T., Long, G., Jiang, J., Pan, S., Zhang, C.: Disan: Directional self-attention network for rnn/cnn-free language understanding. In: Proceedings of the AAAI conference on artificial intelligence. vol.~32 (2018)

\bibitem{sitzmann2020implicit}
Sitzmann, V., Martel, J., Bergman, A., Lindell, D., Wetzstein, G.: Implicit neural representations with periodic activation functions. Advances in neural information processing systems  \textbf{33},  7462--7473 (2020)

\bibitem{strumpler2022implicit}
Str{\"u}mpler, Y., Postels, J., Yang, R., Gool, L.V., Tombari, F.: Implicit neural representations for image compression. In: European Conference on Computer Vision. pp. 74--91. Springer (2022)

\bibitem{tancik2020fourier}
Tancik, M., Srinivasan, P., Mildenhall, B., Fridovich-Keil, S., Raghavan, N., Singhal, U., Ramamoorthi, R., Barron, J., Ng, R.: Fourier features let networks learn high frequency functions in low dimensional domains. Advances in Neural Information Processing Systems  \textbf{33},  7537--7547 (2020)

\bibitem{vaswani2017attention}
Vaswani, A., Shazeer, N., Parmar, N., Uszkoreit, J., Jones, L., Gomez, A.N., Kaiser, {\L}., Polosukhin, I.: Attention is all you need. Advances in neural information processing systems  \textbf{30} (2017)

\bibitem{wang2022predicting}
Wang, H., Tao, G., Ma, J., Jia, S., Chi, L., Yang, H., Zhao, Z., Tao, J.: Predicting the epidemics trend of covid-19 using epidemiological-based generative adversarial networks. IEEE Journal of Selected Topics in Signal Processing  \textbf{16}(2),  276--288 (2022)

\bibitem{xia2023implicit}
Xia, X., Mishne, G., Wang, Y.: Implicit graphon neural representation. In: International Conference on Artificial Intelligence and Statistics. pp. 10619--10634. PMLR (2023)

\bibitem{xu2015show}
Xu, K., Ba, J., Kiros, R., Cho, K., Courville, A., Salakhudinov, R., Zemel, R., Bengio, Y.: Show, attend and tell: Neural image caption generation with visual attention. In: International conference on machine learning. pp. 2048--2057. PMLR (2015)

\bibitem{yang2016hierarchical}
Yang, Z., Yang, D., Dyer, C., He, X., Smola, A., Hovy, E.: Hierarchical attention networks for document classification. In: Proceedings of the 2016 conference of the North American chapter of the association for computational linguistics: human language technologies. pp. 1480--1489 (2016)

\bibitem{yu2015lsun}
Yu, F., Seff, A., Zhang, Y., Song, S., Funkhouser, T., Xiao, J.: Lsun: Construction of a large-scale image dataset using deep learning with humans in the loop. arXiv preprint arXiv:1506.03365  (2015)

\bibitem{yuan2022nerf}
Yuan, Y.J., Sun, Y.T., Lai, Y.K., Ma, Y., Jia, R., Gao, L.: Nerf-editing: geometry editing of neural radiance fields. In: Proceedings of the IEEE/CVF Conference on Computer Vision and Pattern Recognition. pp. 18353--18364 (2022)

\bibitem{yuce2022structured}
Y{\"u}ce, G., Ortiz-Jim{\'e}nez, G., Besbinar, B., Frossard, P.: A structured dictionary perspective on implicit neural representations. In: Proceedings of the IEEE/CVF Conference on Computer Vision and Pattern Recognition. pp. 19228--19238 (2022)

\bibitem{zhang2022implicit}
Zhang, K., Zhu, D., Min, X., Zhai, G.: Implicit neural representation learning for hyperspectral image super-resolution. In: 2022 IEEE International Conference on Multimedia and Expo (ICME). pp.~1--6. IEEE Computer Society (2022)

\bibitem{zhao2018attention}
Zhao, S., Zhang, Z.: Attention-via-attention neural machine translation. In: Proceedings of the AAAI Conference on Artificial Intelligence. vol.~32 (2018)

\bibitem{zhao2019attention}
Zhao, Z., Bao, Z., Zhang, Z., Cummins, N., Wang, H., Schuller, B.: Attention-enhanced connectionist temporal classification for discrete speech emotion recognition  (2019)

\bibitem{zhao2019automatic}
Zhao, Z., Bao, Z., Zhang, Z., Deng, J., Cummins, N., Wang, H., Tao, J., Schuller, B.: Automatic assessment of depression from speech via a hierarchical attention transfer network and attention autoencoders. IEEE Journal of Selected Topics in Signal Processing  \textbf{14}(2),  423--434 (2019)

\bibitem{zheng2021rethinking}
Zheng, J., Ramasinghe, S., Lucey, S.: Rethinking positional encoding. arXiv preprint arXiv:2107.02561  (2021)

\bibitem{zheng2022imface}
Zheng, M., Yang, H., Huang, D., Chen, L.: Imface: A nonlinear 3d morphable face model with implicit neural representations. In: Proceedings of the IEEE/CVF Conference on Computer Vision and Pattern Recognition. pp. 20343--20352 (2022)

\bibitem{zhuang2022filtering}
Zhuang, Y.: Filtering in neural implicit functions. arXiv preprint arXiv:2201.13013  (2022)

\end{thebibliography}
